\documentclass[pdflatex,sn-mathphys-num]{sn-jnl}

\usepackage{graphicx}%
\usepackage{multirow}%
\usepackage{amsmath,amssymb,amsfonts}%
\usepackage{amsthm}%
\usepackage{mathrsfs}%
\usepackage[title]{appendix}%
\usepackage{xcolor}%
\usepackage{textcomp}%
\usepackage{manyfoot}%
\usepackage{booktabs}%
\usepackage{algorithm}%
\usepackage{algorithmicx}%
\usepackage{algpseudocode}%
\usepackage{listings}%
\usepackage{longtable} 
\usepackage{breakurl}
\usepackage{caption}
\usepackage{booktabs}  
\usepackage{url}       
\usepackage{array}
\usepackage{hyperref}
\usepackage{longtable}
\usepackage{makecell}
\usepackage{tcolorbox}
\usepackage{colortbl}
\usepackage{color}
\usepackage{adjustbox}
\usepackage{geometry}
\usepackage{changepage}
\usepackage{threeparttable}
\usepackage[figuresright]{rotating}

\newtcolorbox{myframed}{
    colback=lightgray!20,  
    colframe=white,        
    boxrule=0pt,           
    arc=0pt,               
    left=3pt,              
    right=6pt,             
    top=6pt,               
    bottom=6pt             
}

\theoremstyle{thmstyleone}%
\theoremstyle{thmstyletwo}%

\theoremstyle{thmstylethree}%

\begin{document}
\title[Segment Anything in Pathology Images with Natural Language]{Segment Anything in Pathology Images with Natural Language}


\author[1]{\fnm{Zhixuan} \sur{Chen}}\email{zchenhi@connect.ust.hk}
\equalcont{These authors contributed equally to this work.}

\author[1]{\fnm{Junlin} \sur{Hou}}\email{csejlhou@ust.hk}
\equalcont{These authors contributed equally to this work.}

\author[2]{\fnm{Liqi} \sur{Lin}}\email{linliqi@mail.ustc.edu.cn}

\author[1]{\fnm{Yihui} \sur{Wang}}\email{ywangrm@connect.ust.hk}

\author[1]{\fnm{Yequan} \sur{Bie}}\email{ybie@connect.ust.hk}

\author[1,3]{\fnm{Xi} \sur{Wang}}\email{vancywangxi@ust.hk}

\author[4]{\fnm{Yanning} \sur{Zhou}}\email{amandayzhou@tencent.com}

\author[5]{\fnm{Ronald Cheong Kin} \sur{Chan}}\email{ronaldckchan@cuhk.edu.hk}

\author*[1,6,7,8,9]{\fnm{Hao} \sur{Chen}}\email{jhc@cse.ust.hk}

\affil[1]{\orgdiv{Department of Computer Science and Engineering}, \orgname{The Hong Kong University of Science and Technology}, \orgaddress{\country{Hong Kong SAR, China}}}

\affil[2]{\orgdiv{School of Electronic Engineering and Information Science}, \orgname{University of Science and Technology of China}, \orgaddress{\country{Hefei, China}}}

\affil[3]{\orgdiv{Department of Computer Science and Engineering}, \orgname{The Chinese University of Hong Kong}, \orgaddress{\country{Hong Kong SAR, China}}}

\affil[4]{\orgdiv{Tencent AI Platform Department}, \orgaddress{\country{Shenzhen, China}}}

\affil[5]{\orgdiv{Department of Anatomical and Cellular Pathology}, \orgname{The Chinese University of Hong Kong}, \orgaddress{\country{Hong Kong SAR, China}}}

\affil[6]{\orgdiv{Department of Chemical and Biological Engineering}, \orgname{The Hong Kong University of Science and Technology}, \orgaddress{\country{Hong Kong SAR, China}}}

\affil[7]{\orgdiv{Division of Life Science}, \orgname{The Hong Kong University of Science and Technology}, \orgaddress{\country{Hong Kong SAR, China}}}

\affil[8]{\orgdiv{HKUST Shenzhen-Hong Kong Collaborative Innovation Research Institute}, \orgname{The Hong Kong University of Science and Technology}, \orgaddress{\country{Futian, Shenzhen, China}}}

\affil[9]{\orgdiv{State Key Laboratory of Nervous System Disorders}, \orgname{The Hong Kong University of Science and Technology}, \orgaddress{\country{Hong Kong SAR, China}}}




\abstract{Pathology image segmentation plays a pivotal role in computational pathology, which enables quantitative analysis of histological features for cancer diagnosis and prognosis. However, current segmentation methods encounter significant challenges in clinical applications, primarily due to the scarcity of high-quality, large-scale annotated pathology data and the constraints of fixed, narrowly defined object categories.
To address these issues, this work aims to develop a segmentation foundation model capable of segmenting anything in pathology images using natural language.
First, we establish \textbf{PathSeg}, the largest and most comprehensive dataset for pathology image semantic segmentation, derived from 21 publicly available datasets and comprising 275k image-mask-label triples. Our PathSeg dataset features a wide variety of 160 segmentation categories organized in a three-level hierarchy that covers 20 anatomical regions, 3 histological structures, and 61 object types. 
Next, we introduce \textbf{PathSegmentor}, a text-prompted foundation model tailored for pathology image segmentation. With PathSegmentor, users can achieve semantic segmentation simply by providing a descriptive text prompt for the target category, thus eliminating the need to laboriously provide numerous spatial prompts like boxes or points for each instance. 
Extensive experiments on both internal and external datasets demonstrate the superior segmentation performance of PathSegmentor. 
It outperforms the group of specialized models, effectively handling a broader range of segmentation categories while maintaining a more compact model size.
As a segmentation foundation model, PathSegmentor significantly surpasses other spatial-prompted and text-prompted models by 0.145 and 0.429 improvements in overall Dice scores, respectively, showcasing its remarkable robustness in segmenting complex objects and its effective generalization ability on external evaluations.
Furthermore, we demonstrate that PathSegmentor’s versatile segmentation capabilities can effectively enhance the explainability of classification models for cancer diagnosis through feature importance estimation and imaging biomarker discovery. 
These interpretable outputs provide pathologists with evidence-based decision support, ultimately advancing precision oncology in clinical practice.
}



      
\maketitle

\section{Introduction}\label{sec:introduction}

The segmentation and identification of various structures in pathology images, including tissues, cells, and nuclei, represent a critical foundation for precision medicine \cite{irshad2013methods,wu2022recent}. 
These analyses enable quantitative assessments, such as the calculation of cellular morphology including size, shape, texture, and other imaging features \cite{xing2016robust}.
Recent breakthroughs in deep learning have revolutionized the field of pathology image segmentation, leading to the development of numerous \textbf{specialized models} \cite{chen2017dcan,jiang2023donet,lin2023nuclei,lin2024bonus,graham2019hover,isensee2021nnu,zhang2023sam,horst2024cellvit,horst2025cellvit++} designed for particular tasks and datasets. Although these models often achieve high accuracy, their dependence on predefined segmentation categories limits their effectiveness in real-world clinical settings, which are characterized by diverse and dynamic conditions. 

The emergence of \textbf{segmentation foundation models} presents a transformative solution to address these challenges. 
Segment Anything Model (SAM) \cite{kirillov2023segment} pioneered a novel prompt-driven paradigm that has since been expanded in subsequent segmentation foundation models \cite{ke2023segment,li2024segment,zou2024segment}.
These models enable flexible, interactive segmentation using various prompt types, including spatial prompts (e.g., points, boxes, scribbles), text prompts, reference prompts, or combinations thereof, significantly enhancing their adaptability across diverse applications. This versatility has generated substantial interest in medical imaging analysis.
Several research efforts have explored the medical adaptation of segmentation foundation models \cite{mazurowski2023segment,zhang2024segment}. 
In multimodal image segmentation, models such as MedSAM \cite{ma2024segment}, SAM-Med2D \cite{cheng2023sam}, and BiomedParse \cite{zhao2024biomedparse} have demonstrated effectiveness through distinct prompting strategies, including point/box prompts and text prompts. For specialized imaging modalities, researchers have developed dedicated solutions including $\mu$SAM \cite{archit2025segment} for microscopy analysis, SegVol \cite{du2024segvol} and SAT \cite{zhao2023one} for 3D radiology imaging, and SurgicalSAM \cite{wang2023sam} for endoscopic applications.
Unlike traditional specialized segmentation models, foundation models not only handle a broader range of segmentation tasks, removing the need to train individual models for each application, but also exhibit enhanced generalization capabilities, enabling them to adapt effectively to diverse data distributions in real-world clinical settings.

However, current foundation models for semantic segmentation on pathology images face three significant limitations. 
First, segmentation foundation models designed for multimodal medical images \cite{ma2024segment,cheng2023sam,zhao2024biomedparse} often deliver inferior performance on pathology images because they are trained on a wide range of diverse multimodal medical data, highlighting the urgent need for a domain-specific segmentation foundation model tailored for pathology. 
Second, while a number of foundation models \cite{vorontsov2023virchow,chen2024towards,lu2024visual,ma2024towards,xu2024multimodal} have emerged in pathology for various downstream tasks, foundation models for segmentation tasks still remain markedly underexplored. 
SAM-Path \cite{zhang2023sam} adapts SAM \cite{kirillov2023segment} for pathology segmentation by incorporating trainable class prompts, but transforms it into a specialized model for each dataset. SegAnyPath \cite{wang2024seganypath} is a spatial-prompted foundation model capable of segmenting pathology images across varying resolutions and stain variations. However, spatial-prompted models lack the ability to predict semantic categories and require labor-intensive localization annotations, which become impractical for pathology images with numerous objects. Thus, it is crucial to develop a foundation model that utilizes text prompts to incorporate semantic information, enabling flexible and efficient segmentation.
Moreover, existing frameworks fail to address the inherent semantic ambiguity in pathology image segmentation, where target objects can be interpreted at multiple scales. For example, the term \textit{tumor} may refer to either tissue-level regions or individual cells. To overcome this challenge, it is essential to both construct datasets and develop models that explicitly handle this pathological characteristic by capturing hierarchical semantic information.

In this work, we introduce PathSegmentor, the first text-prompted segmentation foundation model for pathology images.
PathSegmentor efficiently takes pathology images as visual inputs and utilizes hierarchical semantic categories as textual prompts to enable flexible segmentation of diverse objects across various anatomical regions and histological structures. Specifically, our main contributions are:

1) For dataset construction, we curate 275k image-mask-label triples from 21 publicly available pathology image segmentation datasets. Reflecting the inherent complexity of pathology, we reorganize semantic labels hierarchically, yielding 160 categories that encompass 20 anatomical regions, 3 histological structures, and 61 object types. To the best of our knowledge, this dataset, termed PathSeg, stands as the largest and most comprehensive benchmark for semantic segmentation of pathology images. 

2) For model development, we present PathSegmentor, a pathology-specific segmentation foundation model that leverages textual prompts to generate semantic masks for a broad spectrum of pathological categories.
Built on a Transformer encoder-decoder architecture, PathSegmentor incorporates a joint feature interaction module to effectively model the underlying relationships between pathology images and their corresponding categories. This design enables highly flexible, efficient, and semantically aware pathology image segmentation.

3) For experiment evaluation, we conduct extensive experiments in both internal and external validations, and the results demonstrate that:

\begin{itemize}
    \item PathSegmentor demonstrates superior or comparable performance against a group of specialized models (nnUNet \cite{isensee2021nnu}, DeepLabV3+ \cite{chen2018encoder}, and SAM-Path \cite{zhang2023sam}) trained individually for each dataset, while supporting a wider range of segmentation categories with a compact model size.

    \item PathSegmentor achieves significant improvements over spatial-prompted foundation models (MedSAM \cite{ma2024segment} and SAM-Med2D \cite{cheng2023sam}), particularly in segmenting intricate objects with irregular shapes, small sizes, and high density. By leveraging text prompts, PathSegmentor eliminates the need for extensive manual prompting in clinical applications, while delivering semantically aware segmentation results.
    
    \item Compared with the other text-prompted foundation model (BiomedParse \cite{zhao2024biomedparse}), PathSegmentor is trained on large-scale and diverse pathology data with hierarchical semantic labels. This tailored design enables PathSegmentor to offer a more robust solution for pathology image segmentation.

    \item Capitalizing on PathSegmentor's superior segmentation performance, we establish a bidirectional integration with cancer diagnosis classification models for feature importance estimation and imaging biomarker discovery. This greatly enhances the reliability of AI diagnostic models, assisting pathologists in making more informed clinical decisions.

\end{itemize}

\begin{figure}
    \centering
    \includegraphics[width=\linewidth]{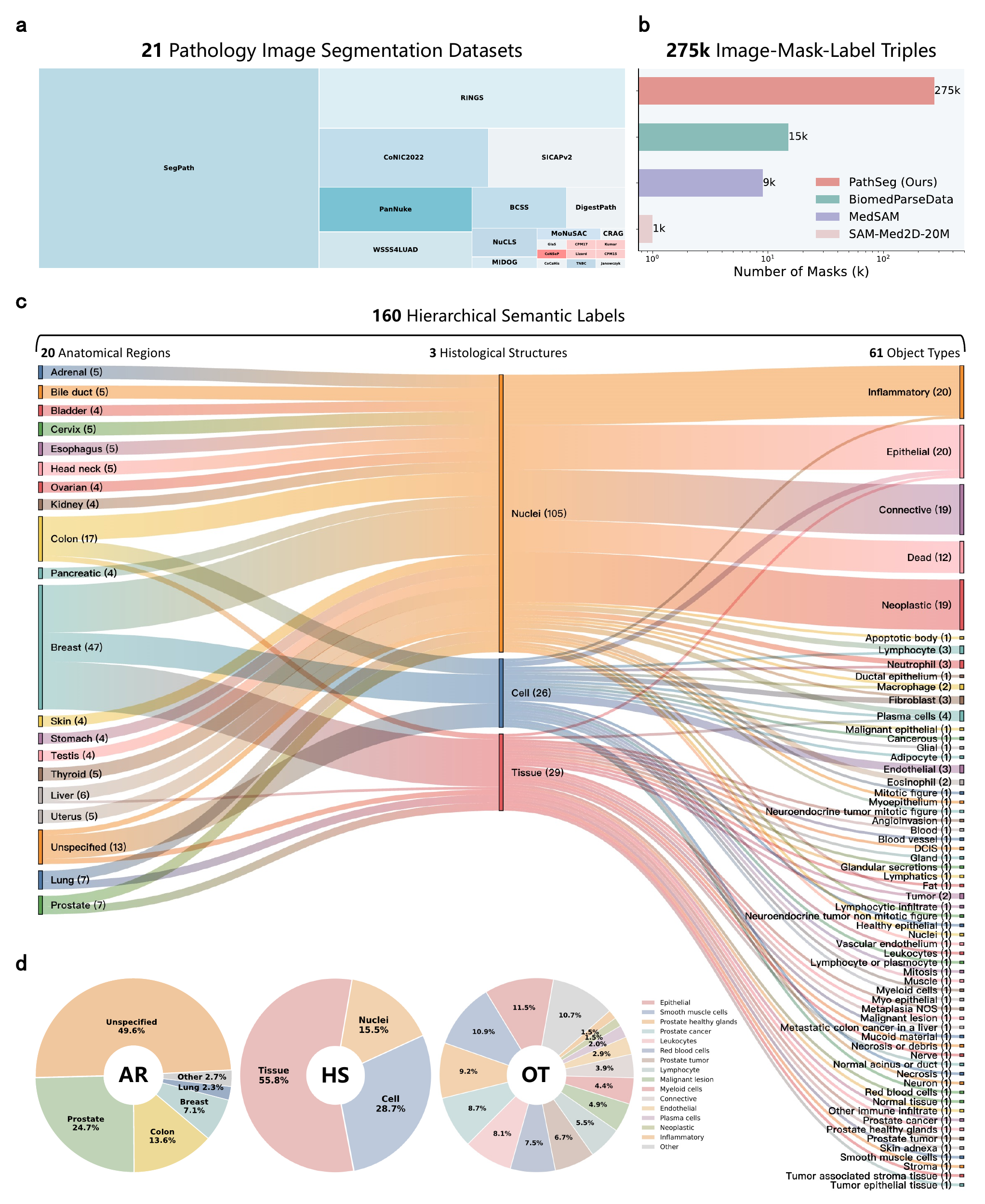}
    \caption{\textbf{Overview of PathSeg dataset.} \textbf{a,} Treemap chart of 21 pathology segmentation datasets in PathSeg (internal, blue; external, red), where area size and color intensity indicate the number of masks and labels, respectively. \textbf{b,} PathSeg contains a total of 275k image-mask-label triples, significantly surpassing the amount of pathology data in other medical image segmentation benchmarks. \textbf{c,} Label distribution in PathSeg illustrated by a Sankey diagram, highlighting the relationships among different anatomical regions (AR), histological structures (HS), and object types (OT). The number in parentheses indicates the count of categories included in each grouping. \textbf{d}, Donut charts visualizing mask distribution across three hierarchical levels in PathSeg.
    }
    \label{fig:pathseg}
\end{figure}

\section{Results}\label{sec:results}

\subsection{Overview of PathSeg Dataset}

Developing a model that can recognize and segment a diverse range of pathological objects requires a large-scale pathology image dataset with comprehensive segmentation annotations. 
However, current public datasets face limitations in both sample size and object type diversity. 
To bridge this gap, we first introduce PathSeg, the largest and most comprehensive dataset for pathology image semantic segmentation, as illustrated in Fig. \ref{fig:pathseg}.

\textbf{Extensive data scale.}
We construct the PathSeg dataset by collecting samples from 21 publicly available datasets (Fig. \ref{fig:pathseg}a) for pathology image semantic segmentation, all containing annotated pixel-level masks with categorical labels.
The complete list of datasets including statistic information and download links is provided in Supplementary Table \ref{tab_supp:datasets_overall}.
The inclusion criteria comprises H\&E-stained histopathology images from human tissue specimens and high-quality segmentation masks with precisely delineated contours. As a result, our PathSeg dataset comprises a total number of 275k image-mask-label triples (Fig. \ref{fig:pathseg}b), which significantly surpasses the pathology subsets of existing medical segmentation datasets such as BiomedParseData (15k) \cite{zhao2024biomedparse}, MedSAM (9k) \cite{ma2024segment}, and SAM-Med2D-20M (1k) \cite{cheng2023sam}. The extensive data and semantic labels in PathSeg dataset provide strong support for the development of effective segmentation foundation models for pathology images.

\textbf{Diverse and hierarchical object categories.}
Different datasets often use varying terminologies or definitions for object types, and merging them directly can lead to labeling inconsistencies. To enhance clarity and standardization, we reorganize the original labels into a unified hierarchical structure, formatted as:
\begin{equation*}
    \text{[anatomical region] - [histological structure] - [object type]},
\end{equation*} 
where [anatomical region] and [histological structure] are extracted directly from official metadata, and [object type] preserve exactly as provided in the source datasets.
This process produces a comprehensive taxonomy of 160 distinct hierarchical semantic labels, systematically organized into three levels (Fig. \ref{fig:pathseg}c):
\begin{itemize}
    \item 20 anatomical regions: such as breast, lung, colon, prostate, etc. \textit{Unspecified} denotes datasets with diverse regional samples without specifying individual image origins. 

    \item 3 histological structures: tissue, cell, and nuclei.
    \item 61 object types: including epithelial, connective, inflammatory, neoplastic, dead, neutrophil, eosinophil, plasma, apoptotic body, lymphocyte, among others.
\end{itemize}
In addition, Fig. \ref{fig:pathseg}c  quantitatively illustrates the hierarchical label distributions, where the breast (47 labels), colon (17), lung (7), and prostate (7) contain the most diverse object types. 
Fig. \ref{fig:pathseg}d presents the mask distributions across three levels. 
Among anatomical regions, the prostate (24.7\%), colon (13.6\%), breast (7.1\%), and lung (2.3\%) exhibit the predominant proportions of masks.
For histological structures, masks are distributed among tissue (55.8\%), cell (28.7\%), and nuclei (15.5\%).
We show the top 15 object types.
The complete numbers of labels, images, and masks for each anatomical region, histological structure, and object type are provided in Supplementary Tables \ref{tab_supp:datasets_anatomical_region}, \ref{tab_supp:datasets_histological_structure}, and \ref{tab_supp:datasets_object_type}, respectively.

\subsection{Overview of PathSegmentor}

\begin{figure}[t]
    \centering
    \includegraphics[width=\linewidth]{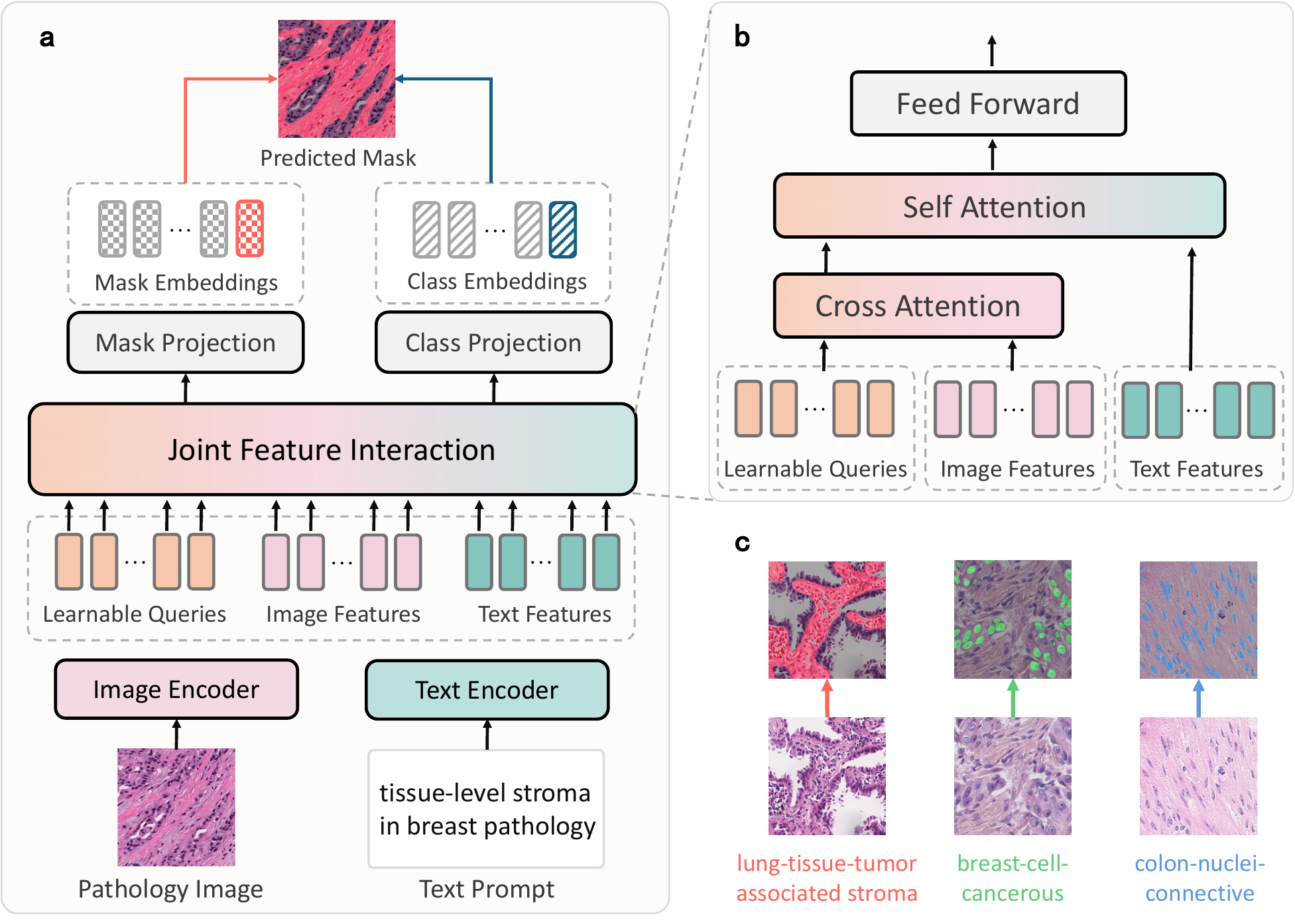}
    \caption{\textbf{Overview of PathSegmentor.} \textbf{a,} The text-prompted framework of PathSegmentor, comprising an image encoder, a text encoder, and a joint feature interaction module. \textbf{b,} The details of the joint feature interaction module, where learnable queries interact with image features and text features through cross-attention and self-attention. \textbf{c,} PathSegmentor can handle diverse segmentation across anatomical regions and histological structures.}
    \label{fig:pathsegmentor}
\end{figure}

As illustrated in Fig. \ref{fig:pathsegmentor}a, PathSegmentor takes a pathology image and a text prompt as inputs, and outputs the predicted mask. Building upon the Transformer encoder-decoder architecture of SEEM~\cite{zou2024segment}, PathSegmentor integrates an image encoder, a text encoder, and a joint feature interaction module with learnable queries. 
First, the image encoder and text encoder transform the pathology image and text prompt into their respective feature representations. Subsequently, the joint feature interaction module combines a set of learnable queries with image and text features to capture the potential association between them through cross-attention and self-attention \cite{zou2023generalized,li2023blip}, resulting in more semantically meaningful representations of mask embeddings and class embeddings. 
Finally, we generate candidate masks via dot-product operations between mask embeddings and image features, then select the final mask based on maximal similarity between class embeddings and text features.

\textbf{Textual prompt generation.}
For the input text, we generate the textual prompts using a predefined template based on the hierarchical semantic labels, enhancing the model's understanding of the context:
\begin{myframed}
\text{[\textbf{histological structure}]-level [\textbf{object type}] in [\textbf{anatomical region}] pathology.}
\end{myframed}

\textbf{Joint feature interaction module.} As one of the key components in PathSegmentor, the joint feature interaction module integrates cross-attention layers, self-attention layers, and feed-forward networks (Fig.~\ref{fig:pathsegmentor}b and Section~\ref{sec:pathsegmentor}).
We create a set of learnable queries, along with image and text features as input to this module. 
Learnable queries first interact with image features through cross-attention, serving as adaptive filters to aggregate critical visual context. 
The vision-enhanced queries are then concatenated with text features and processed via self-attention, facilitating deep interactions where textual semantics guide the refinement of visual features.
The final output from feed-forward networks is used to generate mask embeddings and class embeddings.

PathSegmentor supports versatile pathology image segmentation spanning diverse anatomical regions, histological structures, and object types, as illustrated in Fig.~\ref{fig:pathsegmentor}c. 
All dataset samples are aggregated to train PathSegmentor in an end-to-end manner for pathology image segmentation.
During inference, users simply provide an input pathology image and a text prompt specifying the target using hierarchical semantic information, and the model outputs the corresponding segmentation mask.

\subsection{Segmentation Performance on PathSeg Dataset}

In our experiments, 16 datasets from the PathSeg dataset are partitioned into training and testing splits at a ratio of 8:2, with the testing splits combined for internal validation. We evaluate PathSegmentor against three representative model groups, selected to comprehensively assess performance across different paradigms.
\begin{itemize}
    \item Three specialized segmentation models that are trained and optimized for each dataset independently: following previous works \cite{ma2024segment,zhao2024biomedparse}, we choose 1) nnU-Net \cite{isensee2021nnu}, the most representative medical segmentation model with automated architecture optimization; 2) DeepLabV3+ \cite{chen2018encoder}, which leverages atrous convolution for multi-scale pathology feature extraction; 3) SAM-Path \cite{zhang2023sam}, which implements the frozen SAM encoder \cite{kirillov2023segment} for specialized pathology segmentation.
    \item Two leading spatial-prompted segmentation foundation models adapted from the SAM \cite{kirillov2023segment} architecture for medical imaging: 
    1) MedSAM \cite{ma2024segment}, which improves prompt robustness through 1.5M medical image-mask pairs, specifically optimizing bounding box prompt effectiveness;
    2) SAM-Med2D \cite{cheng2023sam}, which extends the medical data scales to 19.7M while supporting multiple prompts (points, boxes, and masks) for enhanced clinical flexibility.
    \item One text-prompted segmentation foundation model: BiomedParse \cite{zhao2024biomedparse}, representing the state-of-the-art medical model with textual prompting capabilities.
\end{itemize}

\begin{figure}[H]
    \centering
    \includegraphics[width=\linewidth]{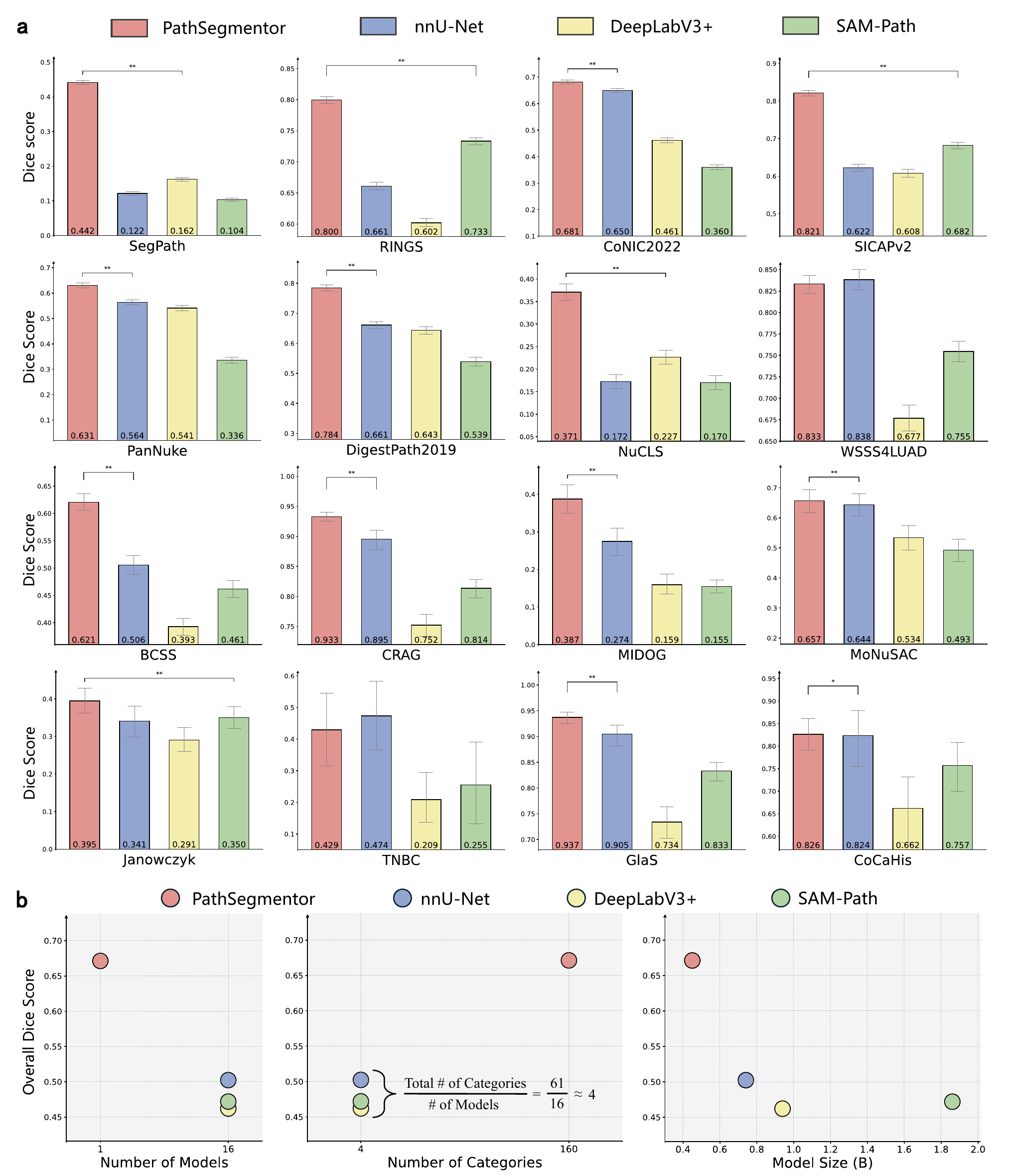}
    \caption{\textbf{Internal validation between PathSegmentor and specialized models across 16 datasets from PathSeg dataset.} \textbf{a}, Bar charts of Dice scores on each dataset, calculated using bootstrapping with 95\% confidence interval. Statistical significance is indicated as **$p\leq 0.01$, *$p\leq0.05$. P-values are computed using one-sided, paired Student’s $t$-tests. \textbf{b}, Scatter plots of overall performance for each model with the number of models, number of categories, and total model size.}
    \label{fig:performance_specific}
\end{figure}

\subsubsection{Comparison with Specialized Segmentation Models}

First, we compare PathSegmentor against three state-of-the-art specialized segmentation models (nnU-Net, DeepLabV3+, SAM-Path) on internal datasets. 
Each specialized model is trained using dataset-optimized configurations through their official implementations.
The quantitative results on 16 datasets are shown in Fig. \ref{fig:performance_specific}a and Supplementary Table \ref{tab_supp:performance_across_datasets}. 
Among the three specialized models, nnU-Net achieves superior Dice scores on 11 out of 16 datasets, while SAM-Path and DeepLabV3+ excel on 3 and 2 datasets, respectively.
Notably, as a segmentation foundation model, PathSegmentor further outperforms all three on 14 out of 16 datasets with statistical significance ($p\leq0.05$, one-sided, paired Student’s $t$-tests). 
Specifically, on SegPath, a challenging multi-category dataset containing eight cell types, PathSegmentor scores 0.442, achieving a 0.280 performance improvement over DeepLabV3+'s score of 0.162. 
The superiority of PathSegmentor can be attributed to large-scale training data and hierarchical semantic modeling. 
Our constructed PathSeg dataset facilitates robust feature learning across diverse pathology patterns, whereas specialized models are constrained by the sizes of individual datasets.
Additionally, hierarchical textual inputs explicitly encode structured pathology knowledge to enhance morphological discrimination.
These results illustrate how foundation models trained on comprehensive data with semantic priors can overcome the limitations of traditional paradigms that rely on one specialized model per dataset.

We also compare the overall performance of four models by averaging the Dice scores of all samples on 16 internal datasets.
As illustrated in Fig. \ref{fig:performance_specific}b, PathSegmentor achieves a superior overall Dice score of 0.671, compared to 0.502 for nnU-Net, 0.462 for DeepLabV3+, and 0.472 for SAM-Path. PathSegmentor exhibits three key advantages.
1) \textbf{Unified architecture}: a single PathSegmentor model outperforms a group of 16 specialized nnU-Net models with a large margin of 0.169 Dice score. 2) \textbf{Category scalability}: PathSegmentor can support all 160 semantic classes in the PathSeg dataset, which is 40$\times$ more than the average capacity of specialized models, each handling approximately 4 classes based on the total of 16 models for 61 object types.
3) \textbf{Model efficiency}: PathSegmentor has a significantly smaller model size of 0.45B parameters and achieves a 75\% size reduction compared to the group of 16 SAM-Path models, which have 1.86B parameters. This makes PathSegmentor more suitable for clinical deployment.
In addition, we present more detailed anatomical region level results in Supplementary Table \ref{tab_supp:performance_across_organs}, histological structure level results in Supplementary Table \ref{tab_supp:performance_across_scales}, and object type level results in Supplementary Table \ref{tab_supp:performance_across_targets}.

\subsubsection{Comparison with Spatial-prompted Segmentation Foundation Models}

In this experiment, we compare PathSegmentor with two state-of-the-art spatial-prompted segmentation foundation models for medical imaging, i.e., MedSAM \cite{ma2024segment} and SAM-Med2D \cite{cheng2023sam}. 
We generate two types of oracle bounding box prompts based on the ground-truth masks.
\begin{itemize}
    \item Instance box: a set of individual boxes, each representing the smallest rectangular bounding box around each instance.
    \item Union box: a single box that serves as the smallest rectangular bounding box encompassing all instance boxes.
\end{itemize}
\noindent To ensure a fair comparison between spatial-prompted models and our text-prompted PathSegmentor, we first use union box prompts for evaluation. These prompts provide a single box for each image, making them comparable to PathSegmentor’s single description per image approach, rather than instance boxes that require dense localization.
Additionally, we will analyze the performance of instance boxes and their prompt efficiency in the intricate objects analysis part.

\begin{figure}[H]
    \centering
    \includegraphics[width=\linewidth]{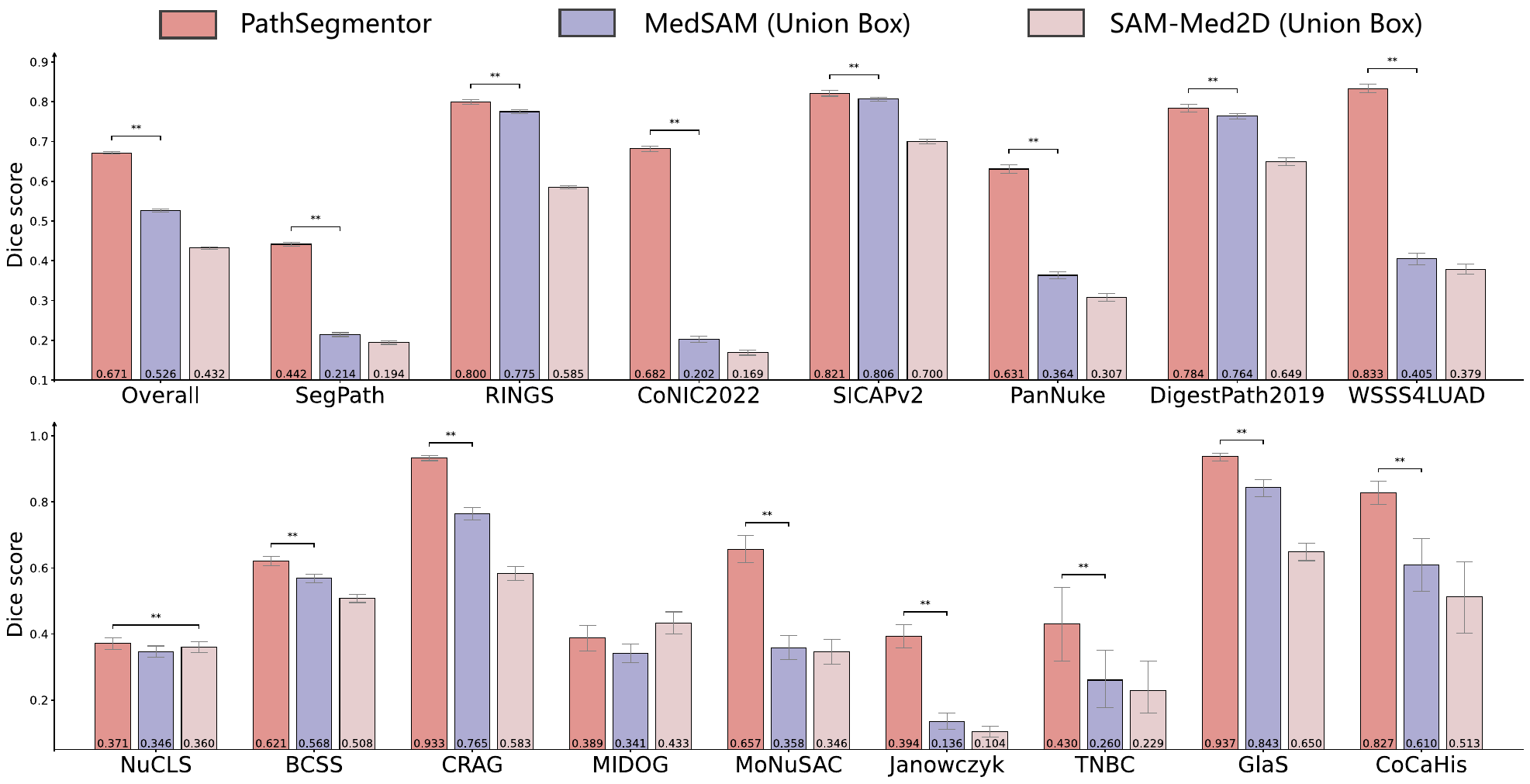}
    \caption{\textbf{Internal validation between PathSegmentor and spatial-prompted segmentation foundation models across 16 datasets from PathSeg dataset.} Bar charts of Dice scores for overall performance and each dataset, calculated using bootstrapping with 95\% confidence interval. Statistical significance is indicated as **$p\leq0.01$. P-values are computed using one-sided, paired Student’s $t$-tests.}
    \label{fig:performance_spatial}
\end{figure}

As illustrated in Fig. \ref{fig:performance_spatial}, PathSegmentor achieves superior overall segmentation performance, outperforming MedSAM by 0.145 and SAM-Med2D by 0.239 in Dice scores.
Across 16 datasets, PathSegmentor shows statistically significant enhancements over MedSAM on 15 of them, with particularly pronounced gains exceeding 0.4 on challenging datasets such as CoNIC2022 and WSSS4LUAD.
These improvements underscore the limitations of MedSAM's general medical pretraining when applied to complex pathology-specific segmentation tasks.
Notably, PathSegmentor maintains superiority even on MedSAM’s training-included GlaS dataset, with a 0.094 Dice score improvement that suggests inherent architectural advantages. While showing a slight 0.044 Dice score disadvantage on MIDOG, PathSegmentor demonstrates robust competitiveness in the internal validation, highlighting text prompting’s efficacy in capturing pathological semantics beyond spatial cues.

\begin{figure}[H]
    \centering
    \includegraphics[width=\linewidth]{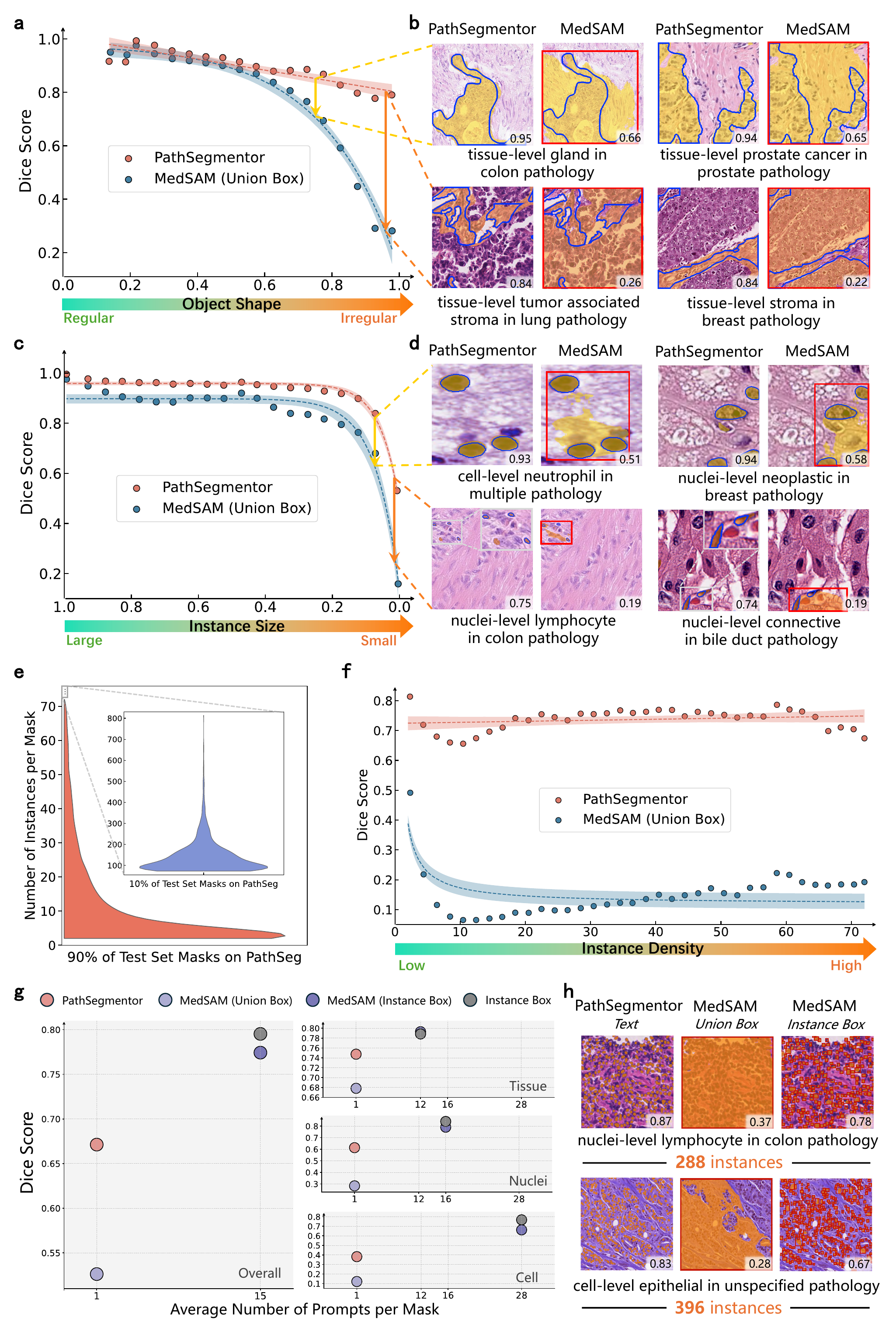}
\end{figure}

\clearpage

\begin{figure}[H]
    \centering
    \caption{\textbf{Comparison of segmentation performance on intricate objects between PathSegmentor and MedSAM.} 
    \textbf{a,} Performance trends for irregular object shapes, with the x-axis representing the metric of irregularity.
    \textbf{b,} Examples of segmentation for irregular tissues. The blue boundary outlines the ground truth, the colored masks indicate the predictions, and the red bounding boxes represent the spatial prompts provided to MedSAM.
    \textbf{c,} Performance trends for small instance sizes, with the x-axis representing the metric of instance size.
    \textbf{d,} Examples of segmentation for small cells and nuclei. 
    \textbf{e,} Distribution of instance counts per object mask in the PathSeg test set. 
    \textbf{f,} Performance trends related to instance density. 
    \textbf{g,} Prompt efficiency, illustrating the trade-off between performance and the number of prompts required per mask. 
    \textbf{h,} Examples of highly dense nuclei and cells, with 288 and 396 instances in a single image.}
    \label{fig:performance_spatial_complex}
\end{figure}

Next, we evaluate PathSegmentor's advantages in segmenting \textbf{intricate objects}, compared to the best-performing spatial-prompted model MedSAM \cite{ma2024segment}. Intricate objects, characterized by complexities such as irregular shapes, small sizes, and high density, present significant challenges for precise segmentation of pathology images. 
Fig. \ref{fig:performance_spatial_complex}a, c, f illustrate how segmentation performance varies with object shape, instance size, and instance density, respectively. Taking object shape (Fig. \ref{fig:performance_spatial_complex}a) as an example, each scatter point represents the mean performance across samples with the corresponding irregularity values. The dashed lines indicate the best-fit curves for these data points with confidence intervals, reflecting the trend in performance variation.
The fitting functions are presented in Supplementary Table \ref{tab_supp:intricate_function}.

\textbf{1) Irregular object shape.}
Pathological objects often exhibit irregular morphologies due to lesions, inflammation, or tumor infiltration. 
To quantify irregularity, we employ the metric derived from circularity \cite{cox1927method} (Section~\ref{sec:metrics}). 
Fig. \ref{fig:performance_spatial_complex}a and Supplementary Table \ref{tab_supp:intricate_performance_irregular} demonstrate that while both MedSAM and PathSegmentor attain Dice scores exceeding 0.9 for regular-shaped objects, their performance diverges significantly with increasing shape irregularity. MedSAM exhibits a parabolic decline in Dice score that drops to 0.282, whereas PathSegmentor sustains stable performance at 0.790 with only marginal degradation. 
This performance gap under maximal irregularity highlights PathSegmentor's superior shape robustness. 
Visual confirmation is provided in Fig. \ref{fig:performance_spatial_complex}b, where PathSegmentor precisely delineates the complex boundaries of irregular tissues across different anatomical regions that challenge the spatial-prompted approaches.

\textbf{2) Small instance size.}
Small instance sizes frequently occur in pathology images, where objects such as small tumors, individual cells, and nuclei are particularly tiny and challenging to differentiate from the surrounding tissue. For each object type, we use the ratio of the average instance scale to the image scale to reflect the instance size. Fig. \ref{fig:performance_spatial_complex}c and Supplementary Table \ref{tab_supp:intricate_performance_tiny} illustrate the trends in model performance as the instance size decreases.
PathSegmentor consistently surpasses MedSAM across different instance sizes, especially exhibiting a remarkable improvement of 0.372 in Dice score for very small instance sizes approaching zero. 
Fig. \ref{fig:performance_spatial_complex}d presents examples of nuclei and cells with small size, demonstrating that PathSegmentor segments small and individual objects effectively, while MedSAM segments a larger area instead.

\textbf{3) High instance density.}
Pathology images frequently exhibit exceptionally high instance densities, posing significant challenges for segmentation tasks. We use the number
of distinct instances within each object mask to indicate instance density.
In Fig. \ref{fig:performance_spatial_complex}e, statistic analysis of our PathSeg test set reveals an instance count distribution: 90\% of masks contain 0$\sim$72 instances, while the upper 10\% contain cases extending beyond this range, reaching a maximum count surpassing 800 instances within a mask.
This distribution underscores the inherent high density in pathology images.
Fig. \ref{fig:performance_spatial_complex}f and Supplementary Table \ref{tab_supp:intricate_performance_dense} illustrate the variation in segmentation results of PathSegmentor and MedSAM as a function of instance density. When instance density rises, MedSAM suffers a more than 0.3 Dice decline, which is attributed to the union box prompts lacking sufficient spatial information to resolve individual instances in crowded regions.
In contrast, PathSegmentor maintains robust performance with a Dice score around 0.7, demonstrating superior capability in segmenting objects with dense instances through text prompts.

\textbf{4) Prompt efficiency analysis.} 
Furthermore, we analyze prompt efficiency by evaluating the trade-off between performance and the number of prompts across different prompting strategies. This is clinically important as the required number of prompts directly correlates with pathologist annotation time. 
Fig. \ref{fig:performance_spatial_complex}g and Supplementary Table \ref{tab_supp:intricate_performance_vs_spatial_prompt} compare four segmentation approaches in terms of their performance and the number of prompts required per mask, including PathSegmentor, MedSAM (Union Box), MedSAM (Instance Box), and Instance Box which directly uses the instance boxes as prediction. 
PathSegmentor demonstrates superior single-prompt efficiency, achieving a 0.145 higher Dice score than MedSAM (Union Box) when limited to one prompt per image. 
The enhanced performance of MedSAM (Instance Box) and Instance Box primarily results from their extensive reliance on precise spatial localization cues. This requires approximately 15$\times$ more prompts per mask overall, specifically 12 for tissues, 16 for nuclei, and 28 for cells. However, this increased demand creates significant challenges for clinical workflows that limit practical adoption.
Fig. \ref{fig:performance_spatial_complex}h visually demonstrates these challenges in extreme cases, where single patches already contains 288 nuclei and 396 cells. When extended to whole-slide images that typically comprise thousands of such patches, the annotation complexity becomes prohibitively expensive, potentially requiring annotation of millions of individual objects.
These results demonstrate PathSegmentor’s advantage in balancing accuracy with practical usability through its efficient text-prompted paradigm for real-world pathology practice.

\begin{figure}[H]
    \centering
    \includegraphics[width=\linewidth]{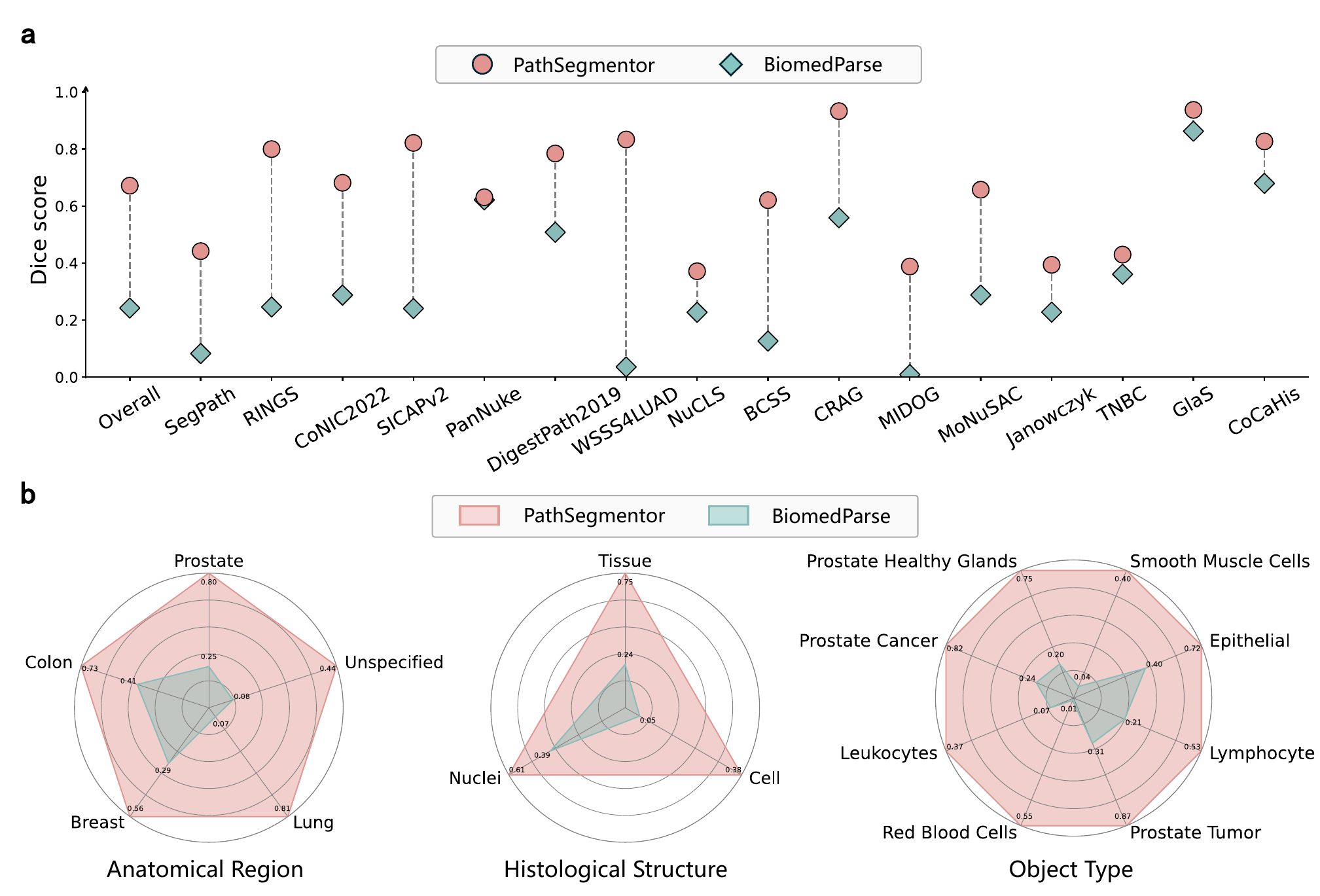}
    \caption{\textbf{Internal validation between PathSegmentor and BiomedParse across 16 datasets from PathSeg dataset.} \textbf{a}, Dice scores for overall performance and individual datasets. \textbf{b}, Results averaged and presented at three levels, including anatomical region, histological structure, and object type.}
    \label{fig:performance_text}
\end{figure}

\subsubsection{Comparison with Text-prompted Segmentation Foundation Models}

In this experiment, we compare PathSegmentor to BiomedParse \cite{zhao2024biomedparse}, a state-of-the-art text-prompted model for biomedical image segmentation. Text-prompted segmentation relies solely on semantic information without incorporating spatial information, making it more challenging. As illustrated in Fig. \ref{fig:performance_text}a and Supplementary Table \ref{tab_supp:performance_across_datasets}, BiomedParse achieves only a 0.242 Dice score for overall performance. It struggles significantly on three complex datasets (SegPath, DigestPath2019, and MIDOG), with Dice scores falling below 0.1. In contrast, after being trained on our extensive and comprehensive PathSeg dataset, PathSegmentor improves Dice score by 0.429 in absolute term over BiomedParse. This substantial enhancement highlights the effectiveness of PathSegmentor. Notably, PathSegmentor consistently exceeds BiomedParse's performance across all 16 datasets, including PanNuke and GlaS, which are also included in BiomedParse's training set, further showcasing its superior performance.

Fig. \ref{fig:performance_text}b presents the comparison results between PathSegmentor and BiomedParse across three specific levels, including anatomical region, histological structure, and object type. For anatomical regions, we focus on five primary areas with over 1,000 masks, including prostate, colon, breast, lung, and unspecified, whereas the results for all regions can be found in Supplementary Table \ref{tab_supp:performance_across_organs}. The results indicate that PathSegmentor demonstrates a significant advantage in segmentation accuracy across these key anatomical regions.
In terms of histological structure, PathSegmentor significantly outperforms BiomedParse across tissue, cell, and nuclei, ranging from large and irregular features to small and dispersed ones (Supplementary Table \ref{tab_supp:performance_across_scales}).
Regarding object type, we showcase the top eight objects with the highest number of masks, including epithelial, smooth muscle cells, prostate healthy glands, prostate cancer, leukocytes, red blood cells, prostate tumor, and lymphocyte, which have mask counts ranging from 15k to 30k. The results for all object types are provided in Supplementary Table \ref{tab_supp:performance_across_targets}. This highlights PathSegmentor's ability to segment a diverse array of biological entities effectively. The segmentation results of all 160 categories in PathSeg dataset are presented in Supplementary Table \ref{tab_supp:performance_across_categories}.

\subsection{External Evaluation}
We assess the generalization performance of PathSegmentor and other segmentation models across pathology images acquired from diverse clinical environments.
The external validation employs datasets that are completely independent from the internal training datasets, while ensuring that the evaluated categories were previously learned during model training.

In our experiments, we conduct external evaluation using 5 datasets where the [object type] in each external category align with categories present in our internal datasets. Specifically, CPM17, CPM15, Kumar, and Lizard datasets contain the nuclei category, while CoNSeP dataset includes 6 types of colon cells.

\begin{figure}
    \centering
    \includegraphics[width=\linewidth]{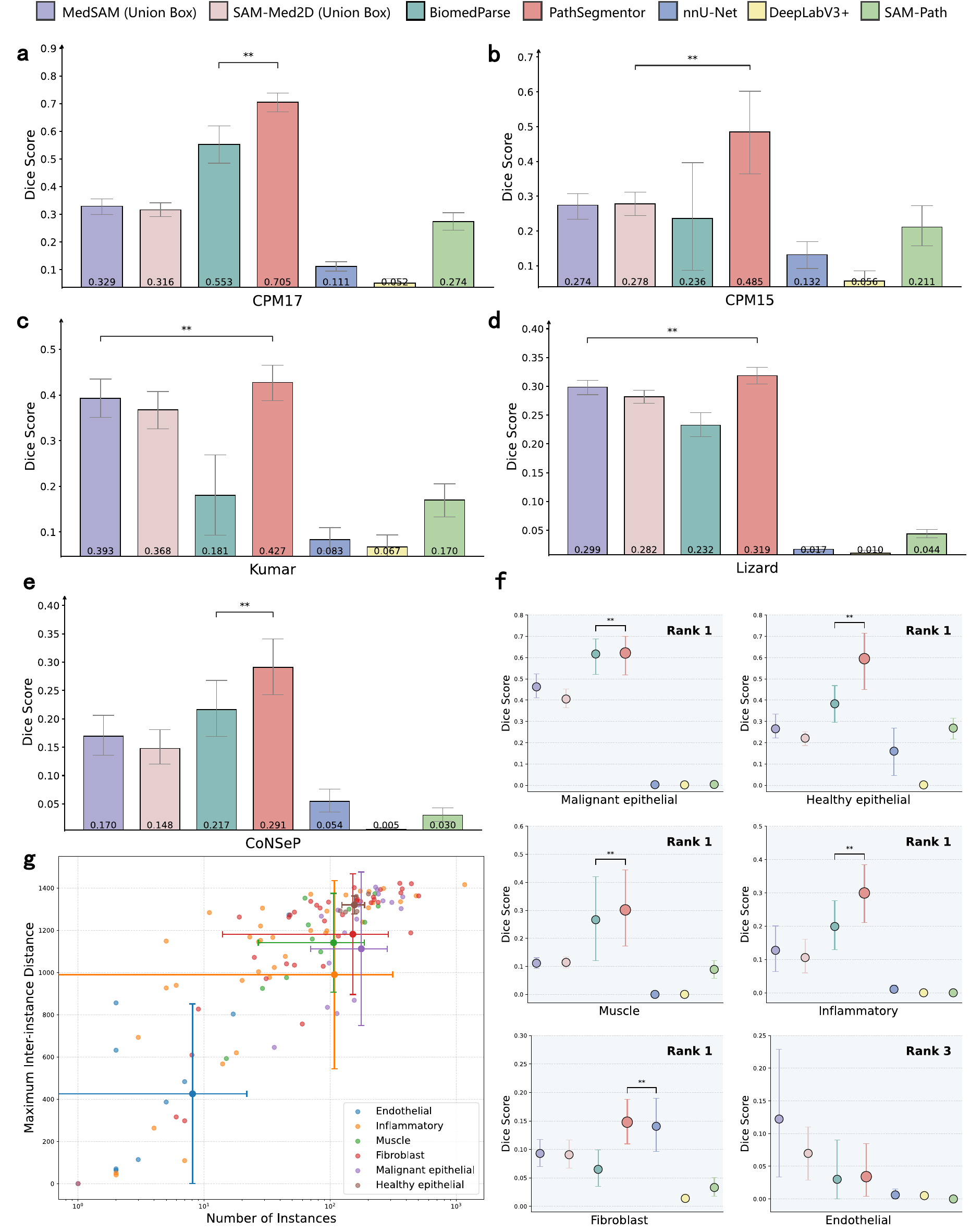}
    \caption{\textbf{External evaluation of PathSegmentor compared to foundation and specialized models.} \textbf{a-e,} Overall Dice scores with bootstrapped 95\% confidence interval on 5 external datasets. Statistical significance is indicated as **$p\leq0.01$. \textbf{f,} Per-category Dice scores for all object types on CoNSeP dataset. PathSegmentor results are visually emphasized through enlarged markers and annotated with rank positions. \textbf{g,} A scatter plot illustrating the number of instances and the maximum inter-instance distance per image on CoNSeP dataset, with horizontal and vertical error bars.}
    \label{fig:external}
\end{figure}

As segmentation foundation models, PathSegmentor, MedSAM, SAM-Med2D, and BiomedParse are directly tested on the external datasets. 
For specialized models, given their tailored configurations for each dataset, we systematically evaluate the specialized models that can handle the corresponding internal category for each external category. As illustrated in Supplementary Table \ref{tab_supp:external_category}, for example, to evaluate the performance on colon-cell-malignant epithelial in CoNSeP, we adopt the specialized models trained on WSSS4LUAD, which involves the category of tumor epithelial tissue. 
For the categories appearing in multiple training datasets, including colon-cell-healthy epithelial, colon-cell-fibroblast, and colon-cell-endothelial, we adopt the results of the best-performing specialized models. 
The external validation results are shown in Fig. \ref{fig:external} and Supplementary Table \ref{tab_supp:performance_external}.

The comparison results of overall performance in Fig. \ref{fig:external}a-e reveal that foundation models (PathSegmentor, MedSAM, SAM-Med2D, and BiomedParse) collectively demonstrate significantly stronger performance than specialized models (nnU-Net, DeepLabV3+, and SAM-Path), highlighting their remarkable generalization capability in real-world clinical settings. 
Among these, PathSegmentor establishes consistent state-of-the-art performance across all external datasets. Specifically for nuclei segmentation, PathSegmentor achieves 0.705, 0.485, 0.427, and 0.319 Dice scores on CPM17, CPM15, Kumar, and Lizard, respectively, demonstrating superior performance in dense object segmentation tasks.
On CoNSeP dataset, PathSegmentor surpasses the second best-performing model BiomedParse by 0.074 Dice score, which clearly shows its enhanced effectiveness.

Furthermore, we focus on the performance of each category in CoNSeP dataset, all of which are colon cells. 
Fig. \ref{fig:external}b illustrates that PathSegmentor with text prompts consistently achieves superior performance for the five cells (malignant epithelial, healthy epithelial, muscle, inflammatory, and fibroblast cells), outperforming all competing models. Particularly, it reaches Dice scores of 0.624 for malignant epithelial and 0.596 for healthy epithelial cells, representing 0.162 and 0.331 improvements over MedSAM (Union Box), the best spatial-prompted model. 
However, we notice an important exception. Although text-prompted models excel for most cell types, both PathSegmentor and BiomedParse underperform spatial-prompted models (MedSAM and SAM-Med2D) on endothelial cells. This divergence motivates our investigation of cellular distribution patterns.
To quantify these differences, we visualize each sample's amount-dispersion distribution in Fig. \ref{fig:external}c, where amount represents the instance count per category in each image, and dispersion measures the maximum inter-instance distance within each category mask. 
Distribution statistics of mean and standard deviation for each category are detailed in Supplementary Table \ref{tab_supp:external_instance_stats}.
The analysis reveals that the five better-performance cell types exhibit both abundant instances and wide spatial distributions, whereas endothelial cells appear sparsely with clustered arrangements. This explains PathSegmentor's strength in segmenting high-density, high-dispersion cellular regions, while spatial prompts provide more precise localization cues for few and concentrated cells.

\subsection{Qualitative Results}

In this experiment, we conduct a comprehensive qualitative analysis of PathSegmentor's segmentation performance through two aspects, including comparative visualizations and practical applications.

\begin{figure}[H]
    \centering
    \includegraphics[width=\linewidth]{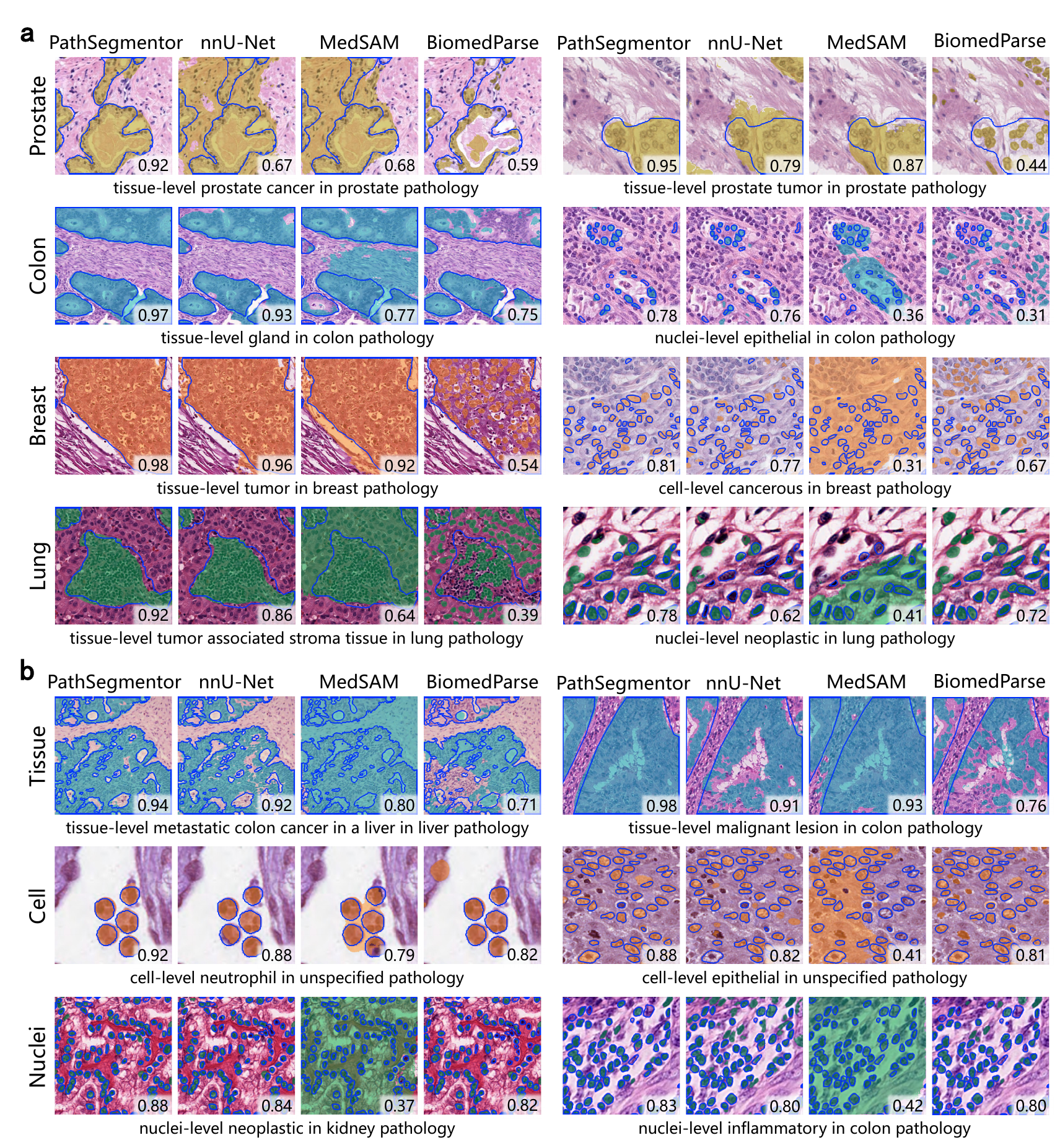}
    \caption{\textbf{Visualization of segmentation results.} We compare PathSegmentor with other state-of-the-art segmentation models, including specialized nnU-Net, spatial-prompted MedSAM, and text-prompted BiomedParse across four anatomical regions and three histological structures. The blue contour lines indicate the ground truth, while the numbers in the bottom right corner of each image represent the Dice score. }
    \label{fig:examples}
\end{figure}

\textbf{Comparative visualizations}. 
Fig. \ref{fig:examples} presents a systematic comparison between PathSegmentor and three representative segmentation models, including specialized nnU-Net, spatial-prompted MedSAM, and text-prompted BiomedParse.
The evaluation spans four critical anatomical regions, including prostate, colon, breast, and lung, along with three histological structures of tissue, cell, and nuclei.
PathSegmentor achieves superior cross-region generalization (Fig. \ref{fig:examples}a), despite significant variability between different anatomical regions in pathological features.
For example, in the prostate images where the target objects exhibit low contrast against surrounding tissues, PathSegmentor can delineate their edges more accurately compared to other models.
Additionally, PathSegmentor showcases its robust multi-scale segmentation capabilities (Fig. \ref{fig:examples}b), effectively handling a range of objects from large tissues such as cancer tissue and malignant lesions, to small and dense nuclei including neoplastic and inflammatory nuclei.
These results highlight PathSegmentor's effectiveness and versatility in segmenting a diverse array of objects across different anatomical regions and histological structures.

\begin{figure}[H]
    \centering
    \includegraphics[width=\linewidth]{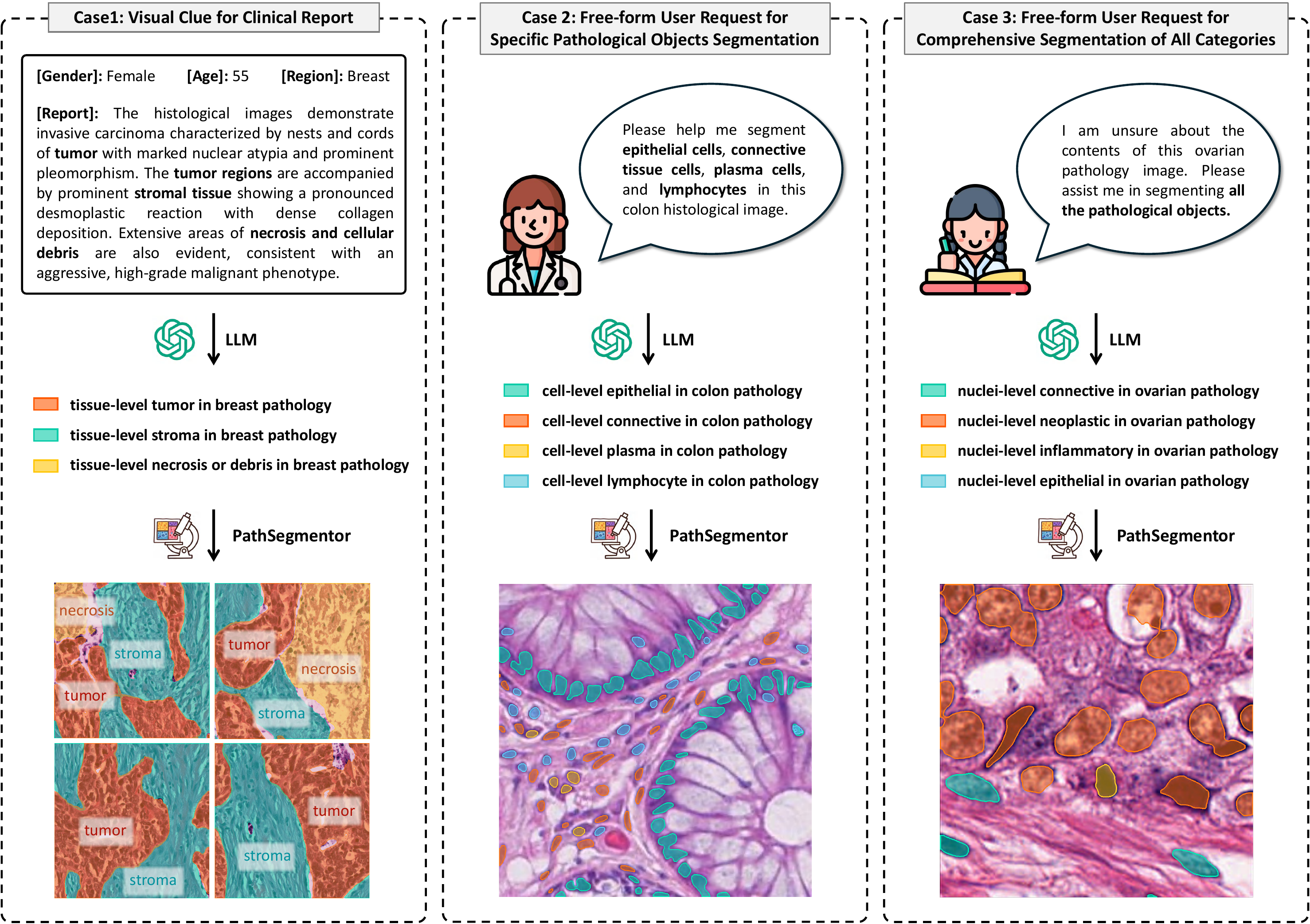}
    \caption{\textbf{Potential applications of PathSegmentor}. PathSegmentor can be seamlessly integrated with LLMs to support interactive segmentation-related tasks, such as providing visual clues for clinical reports, free-form user request for specific pathological objects segmentation or comprehensive segmentation of all categories.
    }
    \label{fig:user_cases}
\end{figure}

\textbf{Potential applications}.
By leveraging its text-prompted segmentation capabilities, PathSegmentor establishes a bridge between natural language and medical image segmentation, serving as an efficient segmentation agent for large language models (LLMs). Fig. \ref{fig:user_cases} illustrates three clinically relevant applications.
\begin{itemize}
    \item Case 1: Report-referenced segmentation, where LLMs automatically parse clinical reports to identify targets of interest and generate optimized textual prompts. These prompts then guide PathSegmentor to produce precise segmentation that visually ground pathological findings, significantly enhancing report interpretation.
    \item Case 2: Targeted conversational segmentation, where LLMs allow end-users to naturally request specific pathological objects segmentation through free-form dialogue, translate these queries into the required prompts in the predefined template for PathSegmentor.
    \item Case 3: Exploratory conversational segmentation, where end-users without prior pathological knowledge can initiate exploratory analysis through open-ended queries. LLMs infer comprehensive segmentation targets based on anatomical regions, prompting PathSegmentor to delineate all relevant objects, which can serve as an interactive educational tool.
\end{itemize}
PathSegmentor's seamless integration with LLMs not only contributes to more flexible segmentation but also significantly improves interactive and versatile medical image analysis. The prompts of LLM used to generate text inputs of target objects for PathSegmentor are detailed in Supplementary Fig. \ref{fig_supp:application_prompt}.

\subsection{PathSegmentor for Explainable Cancer Diagnosis}

Explainable AI in medical image analysis is essential for enhancing the interpretability and trustworthiness of diagnostic models \cite{patricio2023explainable,van2022explainable,hou2024self,hou2024concept}.
In pathology, accurate cancer diagnosis relies on the identification, morphological assessment, and quantitative analysis of specific pathological objects, such as characteristic tissue patterns, cell structures, and nuclear features \cite{diao2021human,kludt2024next}. Therefore, precise segmentation of these objects is critical for diagnostic decision-making.
By extracting and analyzing diagnostically relevant biomarkers, PathSegmentor can generate interpretable, human-understandable explanations that help pathologists make more precise, evidence-based diagnostic judgments.

In this study, we investigate how PathSegmentor's segmentation capability can enhance the interpretability of classification models for cancer diagnosis through two complementary approaches: 1) a \textbf{classification-segmentation} pipeline where we first train a classification model and then employ segmentation to explain the classification model's decisions through object-based feature importance estimation; and 2) a \textbf{segmentation-classification} framework that first identifies pathological objects in pathology images and then performs object-aware classification, facilitating imaging biomarker discovery via object-aware visual explanations.
We evaluate our method using the TCGA-BRCA dataset \cite{TCGA_BRCA} for breast cancer classification, as PathSegmentor can segment the wider variety of objects in breast compared to other anatomical regions. This dataset comprises 985 whole-slide images (WSIs) representing two major breast cancer subtypes, i.e., Invasive Ductal Carcinoma (IDC) and Invasive Lobular Carcinoma (ILC).

\subsubsection{Classification-Segmentation for Feature Importance Estimation}

A common deep learning approach for cancer diagnosis involves training an end-to-end classification model that takes a WSI as input and outputs a diagnostic prediction directly. However, the black-box nature of such models raises concerns about their trustworthiness. Feature importance estimation can provide global explainability of which features are most important to each disease class. This estimation not only verifies whether the model learns the correlation between biologically meaningful patterns and disease classes, but also generates interpretable evidence to assist pathologists by quantifying feature contributions for informed decision-making process. In our experiments, we adopt a classification-segmentation pipeline (Fig. \ref{fig:xai}a), first training a standard cancer diagnosis model based on multi-instance learning \cite{zhou2002neural}, and then using PathSegmentor to segment pathological objects for object-based feature importance estimation to enhance interpretability.

The standard classification model for breast cancer diagnosis is trained on WSIs from the TCGA-BRCA dataset, following the conventional pipeline of patch feature extraction, slide aggregation, and final classification (Fig. \ref{fig:object_model}a and Section~\ref{sec:xai}). This model achieves high classification performance with a macro AUC of 0.936 (Supplementary Table \ref{tab_supp:diagnosis_result}). To investigate model behavior, we implement an object-based perturbation approach to estimate feature importance. 
Different from traditional methods that use non-semantic perturbations, such as gray squares \cite{rise} or superpixel occlusions \cite{ribeiro2016lime}, our approach aims to perform biologically meaningful perturbations. 
We utilize PathSegmentor to segment predefined pathological objects within an image, and then apply localized blurring perturbations exclusively to these segmented regions. Feature importance is quantified by measuring the relative increase in prediction error (Section~\ref{sec:xai}). 
For example, when applying perturbations to segmented regions of breast-tissue-tumor, the model's prediction error increases from 0.003 to 0.012 (Fig. \ref{fig:xai}a, top right). Given a set of predefined pathological objects relevant to IDC and ILC diagnosis, the analysis across all TCGA-BRCA testing samples (Fig. \ref{fig:xai}a, bottom) reveals that these pathological objects generally contribute more significantly to the IDC class than ILC.
Notably, the highly discriminative IDC features discovered by PathSegmentor, including breast-nuclei-ductal epithelium, breast-cell-cancerous, breast-nuclei-neoplastic, breast-tissue-tumor, breast-nuclei-epithelial, and breast-tissue-dcis, well correspond to established pathological diagnostic criteria for the IDC class \cite{rosen2001rosen}. 
These results establish PathSegmentor as an effective explanation tool that both validates model's learning of medically relevant patterns and provides interpretable insights by quantifying decisive features in histological terms.

\begin{figure}[H]
    \centering
    \includegraphics[width=\linewidth]{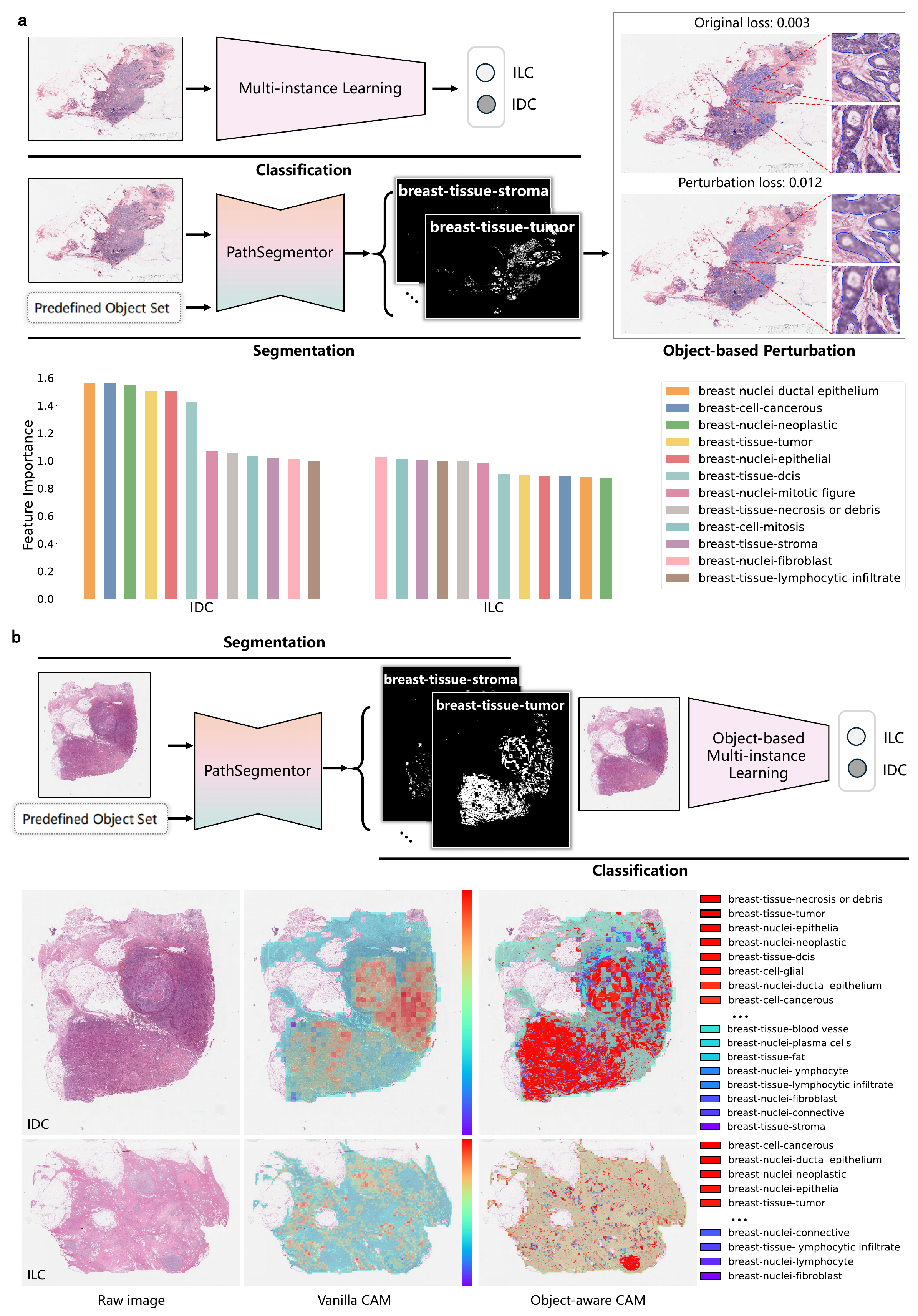}
\end{figure}

\clearpage

\begin{figure}[H]
    \centering
    \caption{\textbf{PathSegmentor supports explainable cancer diagnosis}. \textbf{a}, A classification-segmentation pipeline for object-based feature importance estimation. Features whose perturbation significantly increases model error are identified as important predictors. \textbf{b}, A segmentation-classification framework for imaging biomarker discovery through object-aware class activation maps, which highlight the most discriminative regions with pathological identification. }
    \label{fig:xai}
\end{figure}

\subsubsection{Segmentation-Classification for Imaging Biomarker Discovery}

Another approach to improve the interpretability of standard classification models for cancer diagnosis is to develop an object-aware classification model capable of generating object-aware visual explanations. In our experiments, we build a segmentation-classification framework (Fig. \ref{fig:xai}b), first leveraging PathSegmentor to predict masks of pathological objects, and then decomposing whole-image features into explicit pathological object representations, using object-based multi-instance learning to make final predictions (Fig. \ref{fig:object_model}b and Section~\ref{sec:xai}).
While Class Activation Maps (CAMs) \cite{zhou2016learning} from standard classification models highlight prediction-relevant regions that potentially serve as imaging biomarkers \cite{pai2024foundation},
their clinical utility is limited by an inability to specify the pathological nature of salient regions \cite{rudin2019stop}.
Our object-aware classification model bridges this semantic gap by generating object-aware CAMs, which not only localize discriminative regions but also associate them with specific pathological objects. This capability significantly enhances clinical interpretability, as pathologists require both precise pathological identification and spatial localization to discover biomarkers and inform diagnostic decisions.

After training on the TCGA-BRCA dataset, our object-aware classification model maintains diagnostic performance with a macro AUC of 0.953 for IDC and ILC classification (Supplementary Table \ref{tab_supp:diagnosis_result}). We generate object-aware CAMs by measuring the similarity between object features and classifier weights (Section~\ref{sec:xai}). Comparing vanilla CAMs with object-aware CAMs (Fig. \ref{fig:xai}c), we observe that both methods identify similar discriminative regions indicated by warmer color intensities, while our method additionally provides semantic information about these regions.
For the IDC case in the top row, our object-aware CAM accurately localizes compact tumor mass regions and, importantly, identifies imaging biomarkers with high activation values that include breast-tissue-necrosis or debris, breast-tissue-tumor, breast-nuclei-epithelial.
For the ILC case in the bottom row, our method effectively captures dispersed infiltration patterns of cancerous cells and ranks the salience of pathological biomarkers, highlighting the top three as breast-cell-cancerous, breast-nuclei-ductal epithelium, breast-nuclei-neoplastic. 
By resolving the semantic ambiguity of traditional CAMs, PathSegmentor delivers clinically meaningful visual explanations that assist pathologists in their diagnostic processes and has the potential to uncover previously under-recognized biomarker candidates.

\section{Discussion}
In this study, we aim to develop a foundation model capable of performing flexible segmentation across a wide range of pathological categories. To achieve this, we first construct PathSeg, the largest and most comprehensive pathology image semantic segmentation dataset to date, comprising 275k image-mask-label triples spanning 160 pathological categories. Leveraging the extensive data, we present PathSegmentor, a text-prompted foundational model designed for pathology semantic segmentation.
PathSegmentor stands out from existing segmentation models with the following key advantages:

\textbf{1) One PathSegmentor outperforms a group of specialized models.}
As a segmentation foundation model, it can handle a wide range pathological categories (Fig. \ref{fig:pathseg}c) with a single architecture, eliminating the need for developing a group of specialized models. PathSegmentor achieves comparable or superior performance to specialized models (nnUNet \cite{isensee2021nnu}, DeepLabV3+ \cite{chen2018encoder}, and SAM-Path \cite{zhang2023sam}) across these categories in the internal and external validations (Fig. \ref{fig:performance_specific} and Fig. \ref{fig:external}). 
PathSegmentor's scalability and generalizability introduce a new paradigm, enabling robust segmentation across diverse pathological scenarios.

\textbf{2) Text-prompted PathSegmentor is semantic-aware and user-friendly.} 
PathSegmentor performs flexible semantic segmentation of pathological categories using simple natural language prompts.
Unlike interactive segmentation models, especially those relying on spatial prompts without semantic information (MedSAM \cite{ma2024segment} and SAM-Med2D \cite{cheng2023sam}), text prompts clarify the semantic context of targets, facilitating subsequent analysis and diagnosis. Additionally, text prompts are more user-friendly, enabling segmentation in a single step through semantic information, eliminating the need for multiple interactions (Fig. \ref{fig:performance_spatial_complex}).

\textbf{3) PathSegmentor is a segmentation foundation model uniquely designed for the pathology domain.}
Unlike generic models that attempt to address a broad range of medical image modalities, PathSegmentor is optimized specifically to tackle the complexities of pathological segmentation.
By reformatting all target category labels into a three-level hierarchy of anatomical regions, histological structures, and object types, the model achieves a deeper understanding of pathological features.
This domain-specific design enables it to achieve state-of-the-art performance in segmenting the most comprehensive range of pathology categories (Fig. \ref{fig:performance_text}, compared to BiomedParse \cite{zhao2024biomedparse}), ensuring higher reliability and precision in real-world diagnostic scenarios.

\textbf{4) PathSegmentor drives advancements in explainable pathology.}
PathSegmentor significantly enhances explainable pathology by delivering precise, pixel-level segmentation for accurate measurement of pathological features, which facilitates comprehensive explainability analyses including feature importance estimation and imaging biomarker discovery (Fig. \ref{fig:xai}).
Since PathSegmentor can handle pathology images across a wide range of anatomical regions and pathological categories, it provides accurate explanations for various cancer diagnosis (e.g., breast cancer, lung cancer, prostate cancer), thereby assisting pathologists by offering more reliable decision support in the clinical diagnostic process.

We also highlight the limitations of this work and the potential improvements in future work.

\textbf{1) PathSeg dataset needs to be continuously scaled up.}
While the PathSeg dataset is the largest pathology image dataset for semantic segmentation, its scale remains limited compared to non-semantic datasets \cite{ma2024segment,cheng2023sam,wang2024seganypath}. Scaling up semantic-aware segmentation datasets is resource-intensive in pathology (Fig. \ref{fig:performance_spatial_complex}e). 
We plan to utilize PathSegmentor to implement human-in-the-loop strategies \cite{budd2021survey}, enabling efficient dataset expansion while maintaining high label quality.

\textbf{2) PathSegmentor could benefit from integrating universal prompts.}
PathSegmentor's text prompts enable efficient segmentation; however, performance may decline with semantically novel categories. Spatial prompts provide complementary localization precision. We will combine multiple prompts to enhance robustness \cite{zou2024segment,li2024univs,hou2025qmix}, especially for unseen categories in complex scenarios.

\textbf{3) PathSegmentor requires further real-world clinical validation.}
The clinical utility of PathSegmentor requires validation through multicenter trials that assess real-world segmentation accuracy and robustness. Iterative refinements of PathSegmentor will be guided by pathologist feedback to ensure seamless integration into clinical workflows.

\section{Methods}
\label{sec:methods}
\subsection{PathSeg Dataset Construction}

In this work, we create the PathSeg dataset by integrating 21 publicly available pathology image segmentation datasets, totaling 275k image-mask-label triples. The statistical details and download links of each dataset are presented in Supplementary Table \ref{tab_supp:datasets_overall}.  
The diverse pathology image samples included in the PathSeg dataset cover 20 anatomical regions, 3 histological structures, and 61 object types.

However, integrating these datasets for training presents two key challenges.
1) Ambiguous labels. Original labels typically specify only object types while neglecting critical anatomical and histological context. 
For instance, identically labeled cells from different anatomical regions exhibit distinct morphological characteristics, yet current labels fail to capture these biologically meaningful differences. 
Besides, identical labels may describe distinct biological entities across datasets, such as \textit{tumor} variably denotes either tumor tissue or tumor cells.
This definitional ambiguity critically necessitates standardized label hierarchies that explicitly distinguish between anatomical regions and histological structures.
2) Heterogeneous data. Pathology image datasets exhibit substantial variations in image characteristics, including scales of patch or whole-slide images, image resolutions ranging from $256 \times 256$ to $5412 \times 7215$, and magnification levels of 100$\times$, 40$\times$, 20$\times$, 10$\times$. Such inconsistencies prevent the model from learning unified feature representations.
In the following parts, we present details of label reformat and data standardization.

\textbf{Hierarchical label reformat.}
To avoid potential ambiguity and confusion, we rename the semantic labels in a three-level hierarchical structure, defined as [anatomical region] - [histological structure] - [object type],
where [anatomical region] and [histological structure] are derived from the official metadata, and [object type] preserves the original dataset annotation without additional modification. Note that if a dataset does not specify the anatomical region for each image, the [anatomical region] field should be marked as \textit{unspecified}. 
Cells annotated with region-wise masks instead of individual cell-wise masks, such as plasma cells and smooth muscle cells, are classified under the histological structure category of \textit{tissue}.
As a result, we obtain 20 anatomical regions, 3 histological structures, and 61 object types (statistics detailed in Supplementary Table \ref{tab_supp:datasets_anatomical_region}, Table \ref{tab_supp:datasets_histological_structure} and Table \ref{tab_supp:datasets_object_type}), yielding a total of 160 hierarchical semantic labels (statistics detailed in Supplementary Table \ref{tab_supp:datasets_labels}).

\textbf{Data standardization.}
Prior to model training, we perform data preprocessing to standardize images from different datasets. The processing includes the following steps on both images and corresponding masks. 
1) Magnification normalization. According to dataset metadata, images from 12 datasets in the PathSeg dataset have an initial magnification of $40\times$ (Supplementary Table \ref{tab_supp:datasets_overall}). To standardize magnification, images from the remaining datasets are rescaled to $40\times$ using bilinear interpolation, while the corresponding masks are rescaled using nearest-neighbor interpolation. This approach ensures that the same object type maintains a relatively consistent size across different images, preventing significant variations in appearance.
2) Patching. Whole-slide images are split into a series of image patches using a sliding window. For any image dimension $D$ in height or width, if it exceeds 1500 pixels, we apply a sliding window of size 1024. 
The uniform overlap between adjacent patches is calculated as $\frac{(1024\times \left\lceil D/1024 \right\rceil)-L}{\left\lceil D/1024 \right\rceil - 1}$.
This ensures complete coverage while maintaining spatial regularity. The same method is applied to corresponding masks.
3) Resolution standardization. Image patches are resized to a consistent 1024$\times$1024 resolution via bilinear interpolation, and the mask patches are resized using nearest-neighbor interpolation.  This unified input dimension facilitates batch processing during training.
Finally, we split the datasets for model training and testing (Supplementary Table \ref{tab_supp:datasets_overall}). We use the official split when available. If no official split is provided, we randomly divide the original dataset into 80\% for training and 20\% for testing.

\subsection{PathSegmentor Architecture}
\label{sec:pathsegmentor}
We develop PathSegmentor, a pathology segmentation foundation model that can handle a wide spectrum of pathological objects using textual prompts. 
Formally, each segmentation sample is represented as a triplet $(\textbf{x},\textbf{y},\textbf{t})$, where $\textbf{x}\in \mathbb{R}^{H\times W\times 3}$ is the input pathology image, $\textbf{y}\in \mathbb{R}^{H\times W}$ is the binary mask annotation, $\textbf{t}$ is the textual prompt describing the target object. As depicted in Fig.~\ref{fig:pathsegmentor}a, PathSegmentor follows SEEM \cite{zou2024segment} with three key components, including an image encoder $\Phi_{\text{image}}$ that extracts visual features of the pathology image, a text encoder $\Phi_{\text{text}}$ that 
generates semantic embeddings from the textual prompt, and a joint feature interaction module $\Phi_{\text{joint}}$ where learnable queries fuse multi-modal features through cross-attention and self-attention mechanisms.
Overall, the model predicts the object mask $\mathbf{\hat{y}}$ in the image $\mathbf{x}$ using the textual prompt $\mathbf{t}$:
\begin{equation}
    \mathbf{\hat{y}}=\Phi_{\text{joint}}\left(\Phi_{\text{image}}(\mathbf{x}), \Phi_{\text{text}}(\mathbf{t})\right).
\end{equation}

\textbf{Image encoder}. We employ a FocalNet model~\cite{yang2022focal} as our image encoder $\Phi_{\text{image}}$. For an input pathology image $\mathbf{x}$, the encoder is used to extract visual features represented as a sequence of $m$ tokens with a channel dimension $d$:
\begin{equation}
    F_{\text{image}}=\Phi_{\text{image}}(\mathbf{x})\in\mathbb{R}^{m\times d}. 
\end{equation}

\textbf{Text encoder}. 
The text encoder $\Phi_{\text{text}}$ utilizes a PubMedBERT model~\cite{gu2021domain} to process textual prompts specifying the target object and generate text features of length $L$:

\begin{equation}
    F_{\text{text}}=\Phi_{\text{text}}(\mathbf{t})\in\mathbb{R}^{L\times d}. 
\end{equation}

\textbf{Joint feature interaction module}. 
As illustrated in Fig.~\ref{fig:pathsegmentor}b, we employ a set of learnable queries $\textbf{q}\in\mathbb{R}^{n\times d}$, where $n$ is the number of queries, to effectively extract geometric and semantic information regarding the segmentation target. These queries interact with image features $F_{\text{image}}$ and text features $F_{\text{text}}$ through cross-attention and self-attention layers \cite{zou2023generalized,li2023blip}. 

First, the learnable queries $\textbf{q}\in\mathbb{R}^{n\times d}$ interact with image features $F_{\text{image}}\in\mathbb{R}^{m\times d}$ via a multi-head cross-attention layer. Specifically, with \(\mathbf{q}\) as the query and \(F_{\text{image}}\) as both the key and value, we obtain the image-enhanced queries \(\mathbf{q}'\) as:

\begin{equation}
\begin{aligned}
    \textbf{q}' &= \text{MultiHead-CrossAttn}(\textbf{q}, F_{\text{image}}) \\
    &= \text{Concat}(\text{head}_1, \ldots, \text{head}_h)W_{O_c} \in \mathbb{R}^{n \times d}, \\
    \text{head}_i &= \text{Softmax}\left(\frac{Q_c^{(i)} (K_c^{(i)})^\top}{\sqrt{d_h}}\right)V_c^{(i)} \in \mathbb{R}^{n \times d_h},
\end{aligned}
\end{equation}

\noindent where $Q_c^{(i)} = \textbf{q}W_{Q_c}^{(i)}$, $K_c^{(i)} = F_{\text{image}}W_{K_c}^{(i)}$, and $V_c^{(i)} = F_{\text{image}}W_{V_c}^{(i)}$ are the linear projections for the $i$-th attention head, with projection matrices $W_{Q_c}^{(i)}, W_{K_c}^{(i)}, W_{V_c}^{(i)} \in \mathbb{R}^{d \times d_h}$. The outputs of all heads are concatenated and projected using $W_{O_c} \in \mathbb{R}^{hd_h \times d}$. The term $\sqrt{d_h}$ is a scaling factor to stabilize training.

Second, the image-enhanced queries $\mathbf{q}'\in\mathbb{R}^{n\times d}$ are concatenated with text features $F_{\text{text}}$ to perform multi-head self-attention.
Specifically, $[\mathbf{q}' \,||\, F_{\text{text}}]\in \mathbb{R}^{(n+L)\times d}$ serve as the query, key, and value for computing the joint features $F_\text{joint}$, formulated as:

\begin{equation}
\begin{aligned}
    F_\text{joint} &= \text{MultiHead-SelfAttn}([\mathbf{q}' \,||\, F_{\text{text}}]) \\
    &= \text{Concat}(\text{head}_1, \ldots, \text{head}_h)W_{O_s} \in \mathbb{R}^{(n+L) \times d}, \\
    \text{head}_i &= \text{Softmax}\left(\frac{Q_s^{(i)} (K_s^{(i)})^\top}{\sqrt{d_h}}\right)V_s^{(i)} \in \mathbb{R}^{(n+L) \times d_h},
\end{aligned}
\end{equation}
\noindent where \( Q_s^{(i)} = [\mathbf{q}' \,||\, F_{\text{text}}]W_{Q_s}^{(i)} \), \( K_s^{(i)} = [\mathbf{q}' \,||\, F_{\text{text}}]W_{K_s}^{(i)} \), and \( V_s^{(i)} = [\mathbf{q}' \,||\, F_{\text{text}}]W_{V_s}^{(i)} \). The projection matrices \( W_{Q_s}^{(i)}, W_{K_s}^{(i)}, W_{V_s}^{(i)} \in \mathbb{R}^{d \times d_h} \) are specific to the \(i\)-th attention head, and the final output projection matrix is \( W_{O_s} \in \mathbb{R}^{hd_h \times d} \).

Third, the joint features \( F_\text{joint} \) are passed through a feed-forward network, denoted as \(\text{FFN}\), to obtain enhanced features \( F'_\text{joint} \):
\begin{equation}
    F'_\text{joint} = \text{FFN}(F_\text{joint}) = \text{MLP}(F_\text{joint})\in \mathbb{R}^{(n+L)\times d},
\end{equation}
where \(\text{FFN}\) is composed of a two-layer multilayer perceptron (MLP) with a non-linear activation function between the layers.
The first $n$ tokens of $F'_\text{joint}$ corresponding to the image-enhanced queries $\mathbf{q}'$ are extracted and denoted as the semantic-aware queries $\mathbf{q}''\in\mathbb{R}^{n\times d}$,
which integrate information from both image and text features, enabling semantic segmentation mask generation.

Finally, the semantic-aware queries $\textbf{q}''\in \mathbb{R}^{n\times d}$ are processed through two parallel projection heads, including a mask projector $\mathbf{P}_\text{mask}$ to generate mask embeddings $\mathbf{E}_\text{mask} = \mathbf{P}_\text{mask}(\textbf{q}'')\in\mathbb{R}^{n\times d}$, and a class projector $\mathbf{P}_\text{cls}$ to produce category embeddings $\mathbf{E}_\text{cls} = \mathbf{P}_\text{cls}(\textbf{q}'')\in\mathbb{R}^{n\times d}$. The mask projector $\mathbf{P}_\text{mask}$ is implemented as a three-layer MLP, whereas the class projector $\mathbf{P}_\text{cls}$ is realized using a single fully connected layer. Note that each query in $\textbf{q}''$ generates a corresponding mask embedding and a class embedding. The mask embeddings $\textbf{E}_\text{mask}=\{\textbf{e}_\text{mask}^i\}_{i=1}^n$ are used to decode candidate mask logits, representing the probability of each pixel belonging to the segmentation target. The class embeddings $\textbf{E}_\text{cls}=\{\textbf{e}_\text{cls}^i\}_{i=1}^n$ encode the semantic information of the candidate masks, which can be used to classify their categories.
The final segmentation mask is selected by matching class embeddings with the input text prompt's global embedding $F'_\text{text}$, which corresponds to the last token of $F_\text{text}$:
\begin{equation}
    \hat{\textbf{y}} = \textbf{e}_\text{mask}^j, ~j=\arg\max_{i} \operatorname{Sim}(\textbf{e}_\text{cls}^i,F'_\text{text}),
\end{equation}
where the similarity metric $\operatorname{Sim()}$ is defined as cosine similarity.

\textbf{Objective function.}
We optimize the model using a loss function combining binary cross-entropy (BCE) and Dice loss. For a ground-truth mask $\textbf{y}\in\{0,1\}^{H\times W}$ and predicted mask $\hat{\textbf{y}}\in\{0,1\}^{H\times W}$, the per-sample loss is computed as:
\begin{equation}
\mathcal{L}(\mathbf{y}, \hat{\mathbf{y}}) = \lambda_1\mathcal{L}_{bce} + \lambda_2\mathcal{L}_{dice},
\end{equation}
\begin{equation}
    \mathcal{L}_{bce} = -\frac{1}{HW}\sum_{i,j}\left(y_{ij}\log(\hat{y}_{ij}) + (1-y_{ij})\log(1-\hat{y}_{ij})\right), 
\end{equation}
\begin{equation}
    \mathcal{L}_{dice} = 1 - \frac{2\sum_{i,j} y_{ij}\hat{y}_{ij} + \epsilon}{\sum_{i,j} y_{ij} + \sum_{i,j} \hat{y}_{ij} + \epsilon},
\end{equation}
where $\lambda_1,\lambda_2$ are weighting hyperparameters, $\epsilon$ is used for for numerical stability.

\subsection{Explainable Cancer Diagnosis}
\label{sec:xai}
To demonstrate how the comprehensive segmentation capability of PathSegmentor can support explainable cancer diagnosis, we investigate two analyses, including 
a classification-segmentation pipeline for object-based feature importance estimation and a segmentation-classification framework for imaging biomarker discovery. 

\subsubsection{Breast Cancer Subtyping Dataset}
The experiments are conducted on a breast cancer subtyping dataset derived from the TCGA-BRCA dataset \cite{TCGA_BRCA}, which includes 787 slides of invasive ductal carcinoma (IDC) and 198 slides of invasive lobular carcinoma (ILC). For training and evaluation, the dataset was label-stratified into train-validation-test folds in a ratio of 7:1:2, resulting in 689 slides for training, 99 slides for validation, and 197 slides for testing.

\subsubsection{Classification-Segmentation for Feature Importance Estimation}

\begin{figure}
    \centering
    \includegraphics[width=\linewidth]{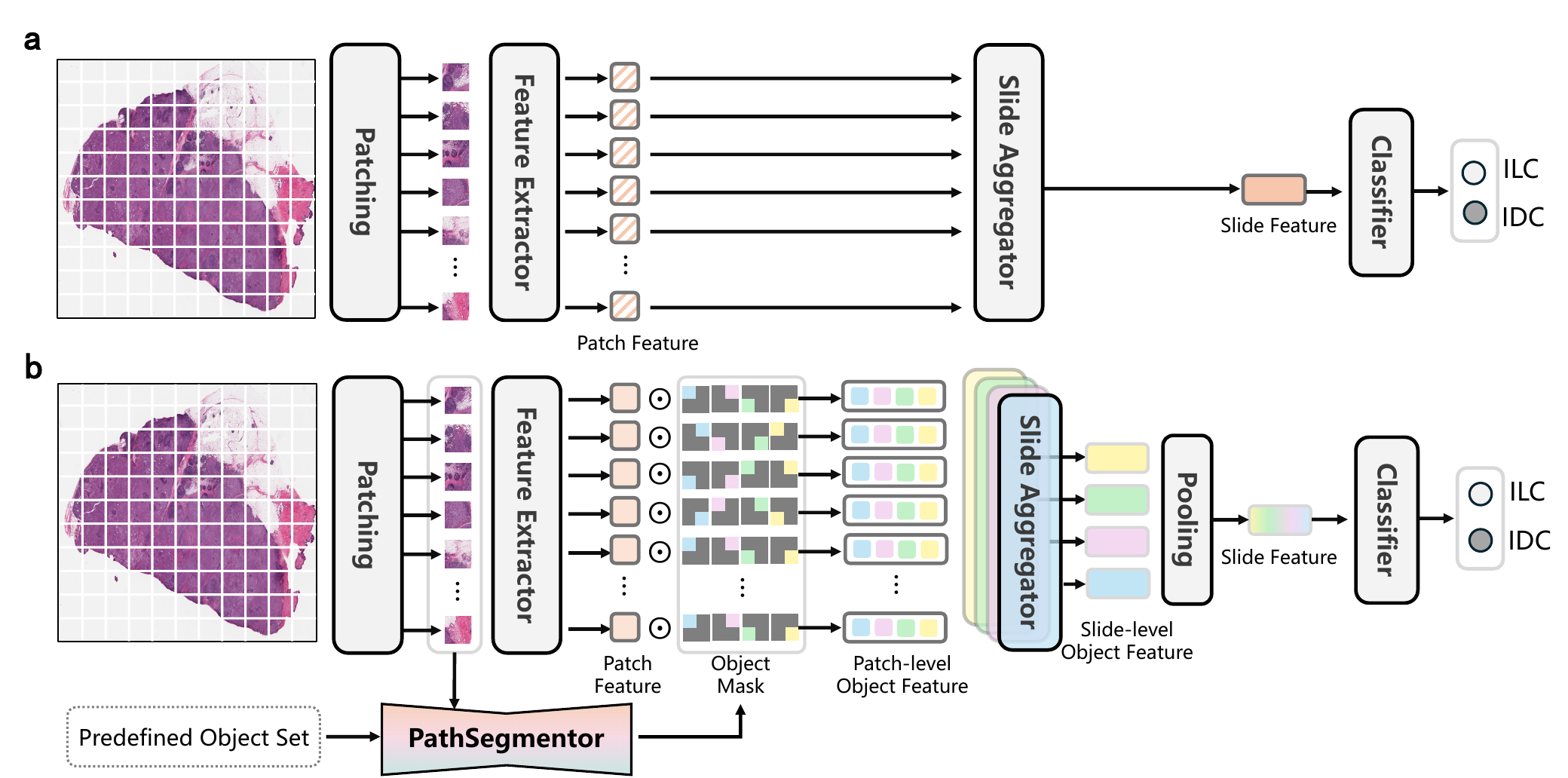}
    \caption{\textbf{Model architectures of the standard model and the object-aware model for cancer diagnosis on WSIs.} \textbf{a}, The standard model consists of patch feature extraction, slide aggregation, and classification. \textbf{b}, The object-aware model incorporates PathSegmentor to decompose patch-features into object-features for object-based visual explanations.
    }
    \label{fig:object_model}
\end{figure}

To enhance the explainability of classification models for cancer diagnosis, we construct a classification-segmentation pipeline. PathSegmentor's segmentation capability can be used to explain a standard classification model through object-based feature importance estimation. First, we train a standard model \cite{gadermayr2024multiple} based on WSIs for cancer diagnosis (Fig. \ref{fig:object_model}a), which involves patch feature extraction, slide aggregation, and diagnostic prediction. In the patch feature extraction step, the input WSI is divided into $N$ non-overlapping patches $X=\{X_i\}_{i=1}^N$. Each patch $X_i\in\mathbb{R}^{H\times W\times 3}$ is fed into a feature extractor $f$ to obtain patch features $P_i=f(X_i)\in \mathbb{R}^{h\times w\times d}$. For slide aggregation, the average pooled patch-level features $P\in\mathbb{R}^{N\times d}$ are then aggregated into a slide-level representation $S=\phi(P)\in \mathbb{R}^{d}$ through a slide aggregator $\phi$. Finally, the slide-level feature $S$ is fed into a classifier $g$ for IDC and ILC classification, resulting in predictions $\hat{Y}=g(S)$. 
The diagnosis model is trained using the standard cross-entropy loss, which minimizes the difference between the prediction $\hat{Y}$ and the ground-truth label $Y$:
\begin{equation}
    \mathcal{L}_{ce}=-(Y\log \hat{Y}+(1-Y)\log(1-\hat{Y})).
\end{equation}

After training the classification model, the importance of each pathological object $o_i$ is calculated as the ratio between the perturbation loss and the original loss, defined as:
\begin{equation}
    \text{IMP}_i = loss_{pert}^i / loss_{orig}.
\end{equation}
Here, the original $loss_{orig}$ denotes the cross-entropy loss produced by the network without any input perturbation. To compute the perturbation loss $loss_{pert}^i$ of the object $o_i$, we first use PathSegmentor to segment the object $o_i$, then apply blurring to its corresponding region in the WSI. This modified image is then fed through the trained classification model to obtain the perturbation cross-entropy loss. The feature importance estimation is conducted on the validation and test set.

\subsubsection{Segmentation-Classification for Imaging Biomarker Discovery}

We build a segmentation-classification framework to enable object-aware visual explanations for imaging biomarker discovery. Specifically, we modify the standard WSI diagnosis model to an object-aware approach through integration with PathSegmentor's segmentation capabilities (Fig. \ref{fig:object_model}b). The modified pipeline consists of four stages, including patch feature extraction, object-aware feature generation, slide aggregation, and diagnostic prediction. 
First, the input WSI is divided into $N$ non-overlapping patches $X=\{X_i\}_{i=1}^N$. Each patch $X_i\in\mathbb{R}^{H\times W\times 3}$ is fed into a feature extractor $f$ to obtain patch features $P_i=f(X_i)\in \mathbb{R}^{h\times w\times d}$. Next, PathSegmentor segments $L$ predefined pathological objects within each patch, resulting in object masks $M_i=\{M_i^j\}_{j=1}^L$, where $M_i^j\in \mathbb{R}^{h\times w}$ is resized to match the same shape of patch features. Patch features are then decomposed into object-specific features $O_i$ through element-wise multiplication with corresponding object masks and masked average pooling:
\begin{equation}
    O_i\in\mathbb{R}^{L\times d}, ~\text{where} ~O_i^j=\text{MaskedAvgPool}(P_i\odot M_i^j)\in\mathbb{R}^{d}.
\end{equation}
Subsequently, parallel aggregators $\phi=\{\phi_j\}_{j=1}^L$, one for each object class, combine patch-level object features $O^j\in\mathbb{R}^{N\times d}$ into slide-level object features $S^j=\phi_j(O^j)\in \mathbb{R}^d$. Finally, all slide-level object features are pooled into a unified slide feature $\bar{S}=\text{AvgPool}(\{S^j\}_{j=1}^L)\in \mathbb{R}^d$ and passed through a classifier $g$ to generate final predictions $\hat{Y}=g(\bar{S})$. We employ the cross-entropy loss to minimize the difference between the predictions $\hat{Y}$ and the ground-truth label $Y$.

\textbf{Patch-based Class Activation Map.} For the standard classification framework, we generate the activation maps that highlight the most discriminative patches in the WSI. Given the averaged patch features $P_i\in\mathbb{R}^{d}$, attention weights $\alpha\in\mathbb{R}^N$ from the MIL module, and classifier weights $W\in\mathbb{R}^{d\times C}$ for $C$ classes. With regard to the predicted class $c$, we calculate the activation value of the $i$-th patch by taking the dot product between the weighted patch features $P_i\alpha_i\in\mathbb{R}^{d}$ and the corresponding weight $W^c\in\mathbb{R}^d$. The resulting activation value can be regarded as part of the final logit for class $c$, and thus directly reflects the patch’s contribution to the classification:
\begin{equation}
    A_i^c=\sum_d W^c_d\cdot P_{i,d}\alpha_i.
\end{equation}
The activation value $A^c(x,y)$ at spatial location $(x,y)$ is defined by: 
\begin{equation}
    A^c(x,y)=A^c_i, ~\text{if}~ (x,y)\in \text{Patch} ~i.
\end{equation}

\textbf{Object-aware Class Activation Map.} From our object-aware classification model, we generate the object-aware CAM that can reflect which specific objects contribute most to the prediction.
Given the slide-level object feature $S^j\in\mathbb{R}^{d}$ and the weights of the classifier $W\in\mathbb{R}^{d\times C}$, we focus on the predicted class $c$ by selecting the corresponding class-specific weight vector $W^c\in\mathbb{R}^d$. Similar to the method used in the patch-based activation map, the activation value of the $j$-th object is computed as the dot product between $S^j$ and $W^c$, which can be viewed as a partial classification logit and quantifies the object's contribution to the prediction result.

\begin{equation}
    A^c_j=\sum_d W^c_d\cdot S^j_d.
\end{equation}
The activation value $A^c(x,y)$ at spatial location $(x,y)$ on the WSI is defined by: 
\begin{equation}
    A^c(x,y)=A^c_j, ~\text{if}~ (x,y)\in \text{Object} ~j.
\end{equation}

\subsection{Implementation Details}
\textbf{Pathology image segmentation.} The input images are resized to $1024 \times 1024$. During the training phase, data augmentations are applied to samples, including 
random resizing via \texttt{ResizeScale} with a scale range of \texttt{[0.9, 1.1]} followed by fixed-size cropping via \texttt{FixedSizeCrop} to a target resolution of $1024 \times 1024$.
PathSegmentor is trained for 20 epochs with a batch size of 32. We use the AdamW \cite{loshchilovdecoupled} optimizer with a base learning rate of 8e-4 and a weight decay of 5e-2. The values of $\lambda_1$ and $\lambda_2$ used in the loss function are both set to 1. PathSegmentor is initialized using BiomedParse's model weights, and all its parameters are trained. PathSegmentor is implemented in PyTorch \cite{paszkepytorch} and runs on four NVIDIA H20 Tensor Core GPUs. 

\noindent \textbf{Explainable cancer diagnosis.} 
For both standard and object-based cancer diagnosis models (Fig. \ref{fig:object_model}), we employ the image encoder of CONCH \cite{lu2024visual} as a frozen feature extractor, training only the slide aggregator and classifier. We adopt Attention-Based Multiple Instance Learning (ABMIL) \cite{ilse2018attention} as the slide aggregator and a fully connected layer as the classifier.
Following the CLAM \cite{lu2021data} toolkit's preprocessing pipeline, we extract patches with 1024$\times$1024 pixels at level 0 for all slides, resizing them to 224$\times$224 pixels while excluding slides with insufficient tissue area. We use the Adam \cite{adam2015} optimizer with a learning rate of 2e-4 and a weight decay of 1e-5. The diagnosis models are implemented in PyTorch \cite{paszkepytorch} and run on one NVIDIA GTX3090 GPU.

\noindent \textbf{Objects used in explainability analysis.} 
From the 47 initially segmented breast tissue objects (Supplementary Table \ref{tab_supp:datasets_anatomical_region}), we apply a Dice score threshold (0.4) to select 21 high-performance pathological objects for imaging biomarker discovery analysis, ensuring reliable segmentation quality for downstream tasks. 
For feature importance estimation, we further refine this set to the 12 objects most diagnostically relevant to IDC/ILC differentiation according to medical literature \cite{rosen2001rosen,christgen2016lobular}. In the implementations of imaging biomarker discovery, we augment these 21 objects with an \textit{other} class encompassing all remaining WSI regions, thereby maintaining comprehensive tissue coverage.
The detailed lists of pathological objects utilized in each experiment are provided in Supplementary Table \ref{tab_supp:xai_object}.

\subsection{Evaluation Metrics}
\label{sec:metrics}
\textbf{Segmentation.} To evaluate the segmentation performance of PathSegmentor and other competing models, we employ the Dice score \cite{milletari2016v}, a standard metric for evaluating medical image segmentation. This score quantifies the overlap between model’s prediction $\hat{Y}$ and ground truth $Y$, and is formally defined as follows:
\begin{equation}
Dice(\hat{Y},Y)=\frac{2|\hat{Y}\cap Y|}{|\hat{Y}|+|Y|}.
\end{equation}

\noindent \textbf{Object characteristics quantitation}. We adopt three quantitative metrics that assess shape irregularity, instance size, and instance density to evaluate intricate object segmentation. For a given ground-truth object mask $M$ in image $I$:

1) \textbf{Shape Irregularity}, which quantifies the deviation of the object’s shape from a perfect circle \cite{cox1927method}, calculated as:
\begin{equation}
    \text{Irregularity} = 1-\frac{4 \pi \cdot \text{Area}(M)}{\left(\text{Perimeter}(M)\right)^2},
\end{equation} 
where $\text{Area}(M)$ represents the total number of pixels within $M$, and $\text{Perimeter}(M)$ denotes the number of pixels along the boundary of $M$. Higher irregularity values indicate greater shape complexity.

2) \textbf{Instance Ratio}, which measures the relative size of individual instances within the object mask $M$, defined as:
\begin{equation}
    \text{InstanceRatio} = \frac{1}{N} \sum_{i=1}^{N} \frac{\text{Area}(\text{Instance}_i)}{\text{Area}(I)},
\end{equation}
where $\text{Area}(\text{Instance}_i)$ and $\text{Area}(I)$ represent the pixel count of the $i$-th instance in $M$ and the entire image $I$, respectively, and $N$ is the total number of instances in $M$. 
Smaller ratios indicate finer instances, which are typically harder to segment accurately.

3) \textbf{Instance Count}, which quantifies instance density by measuring the number of distinct instances within each object mask. To obtain this, we apply connected component analysis using 8-connectivity to identify spatially separated regions within each binary mask~\cite{he2017mask}. Each connected region is treated as an individual instance. To reduce the effect of annotation noise, we filtered out instances smaller than 36 pixels.

4) \textbf{Instance Dispersion},
which quantifies the spatial distribution of object instances within the image $I$, is measured in this work as the maximum pairwise Euclidean distance between instance centroids, defined as:
\begin{equation}
\text{InstanceDis} = \max_{i \ne j} || c_i - c_j ||_2,
\end{equation}
where $c_i$ and $c_j$ denote the centroids of the $i$-th and $j$-th instances, respectively. The centroids are computed using connected component analysis with 8-connectivity on each ground truth mask. To reduce the impact of annotation noise, we exclude instances with fewer than 36 pixels before computing pairwise distances.

\textbf{Cancer diagnosis.} We evaluate our cancer subtyping models using three comprehensive metrics, including Macro AUC, Macro Accuracy (ACC), and Macro F1 score. These metrics are computed by averaging the performance scores across all diagnostic categories, ensuring balanced assessment regardless of class distribution.

\subsection{Competing Methods}
We compare PathSegmentor with three types of state-of-the-art methods, i.e., specialized segmentation models, spatial-prompted segmentation foundation models, and text-prompted segmentation foundation models. 

\textbf{Specialized segmentation models} include nnU-Net~\cite{isensee2021nnu}, DeepLabV3+~\cite{chen2018encoder}, and SAM-Path~\cite{zhang2023sam}. 
In prior medical image segmentation studies \cite{ma2024segment,zhao2024biomedparse}, nnU-Net and DeepLabV3+ have been recognized as representative baselines. 
SAM-Path is a specific pathology segmentation model that integrates a frozen SAM image encoder with a frozen pathology encoder to generate features, then utilizes trainable class prompts for mask prediction on each dataset. 
We train each specialized model individually on all 16 datasets of the PathSeg dataset using optimized configurations. Adopting a multi-class segmentation setting, these models predict all category masks in the dataset simultaneously.
In total, we obtain three groups of 16 specialized models, which serve as competitive methods for comparison with our single PathSegmentor model.

\textbf{Spatial-prompted segmentation foundation models} include MedSAM~\cite{ma2024segment} and SAM-Med2D \cite{cheng2023sam}, which adapt the original SAM model on 1.5M and 19.7M medical image masks, respectively. These models facilitate medical image segmentation across various modalities by employing bounding box prompts.
We utilize their official pretrained implementations for evaluation on PathSeg dataset, using bounding box prompts derived from ground-truth masks.
For the overall evaluation, we adopt the union box as the default setting, which represents the tight rectangular boundary that encloses all instances in the ground truth semantic mask. Each mask is associated with a single union box, allowing for a fair comparison with a single text prompt for the image.
In the analysis of prompt efficiency, we also evaluate the instance box, which closely bounds each individual instance within the ground truth mask. Additionally, we consider using the instance box prompts directly as predictions. Each mask can contain numerous instance boxes.

\textbf{Text-prompted segmentation foundation model} includes BiomedParse \cite{zhao2024biomedparse}, the state-of-the-art foundation model based on textual prompts for medical image segmentation. Employing BiomedParse's official pretrained implementation, we assess its performance on PathSeg dataset. The textual prompts of BiomedParse are generated from its predefined template ``[object type] in [anatomical region] pathology'', which omits critical histological structure details that our PathSegmentor explicitly incorporates.

\subsection*{Data Availability}
All datasets in the PathSeg dataset are publicly available, with a full list of dataset links and statistical details being presented in Supplementary Table~\ref{tab_supp:datasets_overall}.

\subsection*{Code Availability}
The code for PathSegmentor is available at \url{https://anonymous.4open.science/r/PathSegmentor-3166}. For the competing methods, we adopt the official implementations of nnU-Net (\url{https://github.com/MIC-DKFZ/nnUNet}), DeepLabV3+ (\url{https://github.com/VainF/DeepLabV3Plus-Pytorch}), SAM-Path (\url{https://github.com/cvlab-stonybrook/SAMPath}), MedSAM (\url{https://github.com/bowang-lab/MedSAM}), SAM-Med2D (\url{https://github.com/OpenGVLab/SAM-Med2D}), and BiomedParse (\url{https://github.com/microsoft/BiomedParse}).

\subsection*{Ethics Declarations}
This project has been reviewed and approved by the Human and Artefacts Research Ethics Committee (HAREC) of The Hong Kong University of Science and Technology. The protocol number is HREP-2025-0024.

\subsection*{Author Contribution}
Z.C., J.H., and H.C. conceived and designed the work. Z.C., L.L., and J.H. curated the data included in the PathSeg benchmark. Z.C. contributed to the technical implementation of the PathSegmentor framework. Z.C. evaluated the performance of segmentation foundation models. L.L. and Z.C. evaluated the performance of task-specific segmentation models. J.H., Z.C., Y.W., and Y.B. designed and conducted the disease diagnosis and biomarker discovery tasks. Z.C. and J.H. wrote the manuscript with inputs from all authors. All authors reviewed and approved the final paper. H.C. supervised the research.

\subsection*{Declarations}
The authors have no conflicts of interest to declare.

\subsection*{Acknowledgment}
This work was supported by the National Natural Science Foundation of China (No. 62202403), Hong Kong Innovation and Technology Commission (Project No. MHP/002/22 and ITCPD/17-9), Shenzhen Science and Technology Innovation Committee Fund (Project No. KCXFZ20230731094059008) and Research Grants Council of the Hong Kong Special Administrative Region, China (Project No. R6003-22 and C4024-22GF).

\bibliography{sn-bibliography}

\clearpage
\begin{appendices}

\section{Supplementary Material}

\begin{table*}[h]
\centering
\caption{An overview of the 21 selected datasets in our PathSeg dataset for pathology image segmentation. The table lists dataset name, magnification (MA), anatomical regions (AR), number of labels, images, masks (train/test masks). }
\label{tab_supp:datasets_overall}
\resizebox{\textwidth}{!}{
\begin{tabular}{lcccccc}
\toprule
\textbf{Data Source} & \textbf{MA} & \textbf{AR} & \textbf{\#Label} & \textbf{\#Image} & \textbf{\#Mask (\#Train/\#Test)} \\
\midrule
\multicolumn{7}{l}{\textit{\textcolor{gray}{Internal Datasets (16)}}}\\
\href{https://github.com/PathologyDataScience/BCSS}{BCSS} \cite{amgad2019structured} & 40x & Breast & 18 & 2,914 & 9,192 (7,322/1,870) \\
\href{https://cocahis.irb.hr/}{CoCaHis} \cite{sitnik2021dataset} & 40x & Liver & 1 & 82 & 82 (58/24)  \\
\href{https://conic-challenge.grand-challenge.org/}{CoNIC2022} \cite{graham2024conic} & 20x & Colon & 6 & 4,841 & 19,332 (15,387/3,945)\\
\href{https://warwick.ac.uk/fac/cross_fac/tia/data/mildnet/}{CRAG}~\cite{graham2019mild} & 20x & Colon & 1 & 1,750 & 1,750 (1,429/321) \\
\href{https://digestpath2019.grand-challenge.org/}{DigestPath2019}~\cite{da2022digestpath} & 40x & Colon & 1 & 13,391 & 13,391 (10,725/2,666) \\
\href{https://datasetninja.com/gland-segmentation\#download}{GlaS}~\cite{sirinukunwattana2017gland} & 20x & Colon & 1 & 165 & 165 (85/80) \\
\bottomrule 
\end{tabular}

}
\end{table*}

\begin{table*}[h]
\centering
\caption{An overview of 20 anatomical regions in PathSeg. For each anatomical region (AR), this table lists numbers of histological structures (HS), object types (OT), labels, images, and masks. }
\label{tab_supp:datasets_anatomical_region}
\begin{tabular}{lccccccccc}
\toprule
\textbf{AR} & \textbf{\#HS} & \textbf{\#OT} & \textbf{\#Label} & \textbf{\#Image} & \textbf{\#Mask}\\
\midrule
Breast & 3 & 44 & 47 & 7,411 & 19,640\\
Lung & 2 & 7 & 7 & 3,043 & 6,182\\
Colon & 3 & 16 & 17 & 21,487 & 37,507\\
Prostate & 2 & 7 & 7 & 62,329 & 67,991\\
Unspecified & 3 & 12 & 13 & 135,486 & 136,089\\
Adrenal & 1 & 5 & 5 & 436 & 863\\
Bile duct & 1 & 5 & 5 & 395 & 861\\
Esophagus & 1 & 5 & 5 & 422 & 729\\
Head neck & 1 & 5 & 5 & 327 & 627\\
Liver & 2 & 6 & 6 & 305 & 566\\
Cervix & 1 & 5 & 5 & 259 & 536\\
Thyroid & 1 & 5 & 5 & 221 & 510\\
Testis & 1 & 4 & 4 & 189 & 428\\
Pancreatic & 1 & 4 & 4 & 195 & 404\\
Uterus & 1 & 5 & 5 & 186 & 395\\
Stomach & 1 & 4 & 4 & 146 & 382\\
Skin & 1 & 4 & 4 & 179 & 373\\
Ovarian & 1 & 4 & 4 & 146 & 290\\
Kidney & 1 & 4 & 4 & 127 & 247\\
Bladder & 1 & 4 & 4 & 130 & 233\\
\bottomrule
\end{tabular}
\end{table*}

\begin{table*}[h]
\centering
\caption{An overview of 3 histological structures in PathSeg. For each histological structure (HS), this table lists numbers of anatomical regions (AR), object types (OT), labels, images, and masks. }
\label{tab_supp:datasets_histological_structure}
\begin{tabular}{lccccccc}
\toprule
\textbf{HS} & \textbf{\#AR} & \textbf{\#OT} & \textbf{\#Label} & \textbf{\#Image} & \textbf{\#Mask}\\
\midrule
Nuclei & 6 & 29 & 105 & 15,993 & 42,300\\
Cell & 3 & 22 & 26 & 78,402 & 78,867\\
Tissue & 20 & 21 & 29 & 139,024 & 153,686\\
\bottomrule
\end{tabular}
\end{table*}

\begin{table*}[h]
\centering
\caption{An overview of 61 object types in PathSeg. For each object type (OT), this table lists numbers of anatomical regions (AR), histological structures (HS), labels, and masks (images). }
\label{tab_supp:datasets_object_type}
\resizebox{\textwidth}{!}{
\begin{tabular}{lccccc}
\toprule
\textbf{OT} & \textbf{\#AR} & \textbf{\#HS} & \textbf{\#Label} & \textbf{\#Mask (Image)}\\
\midrule
Endothelial & 3 & 1 & 3 & 8,118 \\
Epithelial & 18 & 3 & 20 & 31,551 \\
Leukocytes & 1 & 1 & 1 & 22,318 \\
Lymphocyte & 3 & 2 & 3 & 15,267 \\
Macrophage & 2 & 2 & 2 & 312 \\
Myeloid cells & 1 & 1 & 1 & 11,977 \\
Neutrophil & 3 & 2 & 3 & 1,358 \\
\bottomrule
\end{tabular}
}
\end{table*}

\clearpage

\begin{longtable}{
    p{10cm}                                     
    >{\centering\arraybackslash}p{3cm}         
}
\caption{An overview of 160 hierarchical semantic labels in PathSeg. For each label formatted as [anatomical region]-[histological structure]-[object type], this table lists numbers of masks (images). }\\
\toprule
\textbf{Hierarchical Semantic Label} & \textbf{\#Mask (Image)}\\
\midrule
\endfirsthead
\multicolumn{2}{c}{{\bfseries \tablename\ \thetable{} -- continued from previous page}} \\
 \toprule

\textbf{Hierarchical Semantic Label} & \textbf{\#Mask (Image)} \\
 \midrule
 \endhead
 \midrule
 \multicolumn{2}{r}{Continued on next page} \\
 \endfoot
 \bottomrule
 \endlastfoot
Unspecified-Cell-Endothelial & 8,088 \\
Unspecified-Cell-Epithelial & 179 \\
Unspecified-Cell-Leukocytes & 22,318 \\
Unspecified-Cell-Lymphocyte & 9,942 \\
Unspecified-Cell-Macrophage & 84 \\
Unspecified-Cell-Myeloid cells & 11,977 \\
Unspecified-Cell-Neutrophil & 137 \\
Unspecified-Cell-Plasma cells & 5,310 \\
\end{longtable}

\begin{table*}[h]
\centering
\caption{Dice scores for segmentation results across different datasets on the internal datasets of PathSeg. Overall results are reported by averaging the Dice scores of all samples on the internal datasets.
\textbf{Bold} indicates the best result and \underline{underline} indicates the second best. 95\% confidence interval (CI) is included in square brackets. P-values are computed using one-sided, paired Student’s $t$-tests between PathSegmentor and the second-best models.}
\resizebox{\textwidth}{!}{%
\begin{tabular}{lcccccccc}
\toprule
\multirow{2}{*}[-2.ex]{\textbf{Dataset}} & \multicolumn{3}{c}{\textbf{Specialized Model}} & \multicolumn{4}{c}{\textbf{Segmentation Foundation Model}} & \multirow{2}{*}[-2.ex]{\textbf{\textit{P}-value}}\\
            \cmidrule(lr){2-4} \cmidrule(lr){5-8}
     & \textbf{DeepLabV3+} & \textbf{SAM-Path} & \textbf{nnU-Net} & \makecell{\textbf{MedSAM} \\ \textbf{(Union Box)}} & \makecell{\textbf{SAM-Med2D} \\ \textbf{(Union Box)}} & \textbf{BiomedParse} & \textbf{PathSegmentor} \\
\midrule
GlaS & \makecell{0.735 \\ \scriptsize{[0.702, 0.765]}} & \makecell{0.833 \\ \scriptsize{[0.816, 0.850]}} & \makecell{\underline{0.905} \\ \scriptsize{[0.885, 0.922]}} & \makecell{0.843 \\ \scriptsize{[0.814, 0.868]}} & \makecell{0.650 \\ \scriptsize{[0.621, 0.675]}} & \makecell{0.862 \\ \scriptsize{[0.842, 0.880]}} & \makecell{\textbf{0.937} \\ \scriptsize{[0.924, 0.947]}} & $\leq 0.01$ \\
CRAG & \makecell{0.752 \\ \scriptsize{[0.733, 0.772]}} & \makecell{0.814 \\ \scriptsize{[0.798, 0.828]}} & \makecell{\underline{0.895} \\ \scriptsize{[0.878, 0.911]}} & \makecell{0.765 \\ \scriptsize{[0.746, 0.783]}} & \makecell{0.583 \\ \scriptsize{[0.561, 0.604]}} & \makecell{0.559 \\ \scriptsize{[0.531, 0.588]}} & \makecell{\textbf{0.933} \\ \scriptsize{[0.925, 0.940]}} & $\leq 0.01$ \\
WSSS4LUAD & \makecell{0.677 \\ \scriptsize{[0.661, 0.692]}} & \makecell{0.755 \\ \scriptsize{[0.742, 0.767]}} & \makecell{\textbf{0.838} \\ \scriptsize{[0.825, 0.850]}} & \makecell{0.405 \\ \scriptsize{[0.391, 0.419]}} & \makecell{0.379 \\ \scriptsize{[0.365, 0.392]}} & \makecell{0.036 \\ \scriptsize{[0.032, 0.039]}} & \makecell{\underline{0.833} \\ \scriptsize{[0.822, 0.843]}} & - \\
\bottomrule
\end{tabular}
}
\label{tab_supp:performance_across_datasets}
\end{table*}

\begin{table}[h]
\centering
\caption{The fitting functions of the performance trend with regard to the changes on object shape, instance size, and instance density in the intricate object analysis. }
\label{tab_supp:intricate_function}
\begin{tabular}{lcccc}
\toprule
\textbf{Method} & \textbf{Function Type} & \textbf{Formula} & $\mathbf{MAE}$ & $\mathbf{RSME}$ \\
\midrule
\multicolumn{4}{l}{\textit{\textcolor{gray}{Object Shape}}}\\
MedSAM & Exponential & $y=-0.8436\exp(-4.4946x)+0.9802$ & 0.0217 & 0.0305\\
PathSegmentor & Linear & $y=0.2085x + 0.7998$ & 0.0231 & 0.0279\\
\midrule
\multicolumn{4}{l}{\textit{\textcolor{gray}{Instance Size}}}\\
MedSAM & Exponential & $y=-0.7653\exp(-14.1092x)+0.8978$ & 0.0258 & 0.0343\\
PathSegmentor & Exponential & $y=-0.4800\exp(-17.2893x)+0.9586$ & 0.0103 & 0.0135\\ 
\midrule
\multicolumn{4}{l}{\textit{\textcolor{gray}{Instance Density}}}\\
MedSAM & Reciprocal & $y=93.4330/173.6386x + 0.1192$ & 0.0512 & 0.0609\\
PathSegmentor & Linear & $y=0.0003x + 0.7240$ & 0.0280 & 0.0355\\
\bottomrule
\end{tabular}
\end{table}
\begin{table*}[h]
\centering
\caption{Dice results of MedSAM (Union Box) and PathSegmentor on intricate objects with increasing irregularity. The column of Irregularity Interval indicates the interval partitioning of object irregularity. Avg. Irregularity and Avg. Dice represent the weighted average values within each interval, where the weights are determined by Gaussian Kernel Density Estimation based on the density of data points.}
\label{tab_supp:intricate_performance_irregular}
\resizebox{\textwidth}{!}{%
\begin{tabular}{lcccc}
\toprule
\multirow{2}{*}[-0.8ex]{\makecell{\textbf{Irregularity}\\ \textbf{Interval}}} & \multicolumn{2}{c}{\textbf{MedSAM (Union Box)}} & \multicolumn{2}{c}{\textbf{PathSegmentor}} \\
\cmidrule(lr){2-3}\cmidrule(lr){4-5}
& Avg. Irregularity & Avg. Dice & Avg. Irregularity & Avg. Dice \\
\midrule
$[0.10, 0.15)$ & 0.141 & 0.950 & 0.137 & 0.916 \\
$[0.15, 0.20)$ & 0.190 & 0.941 & 0.186 & 0.914 \\
$[0.20, 0.25)$ & 0.218 & 0.975 & 0.218 & 0.993 \\
$[0.25, 0.30)$ & 0.272 & 0.944 & 0.272 & 0.976 \\
$[0.30, 0.35)$ & 0.325 & 0.926 & 0.324 & 0.956 \\
$[0.35, 0.40)$ & 0.376 & 0.919 & 0.375 & 0.944 \\
$[0.40, 0.45)$ & 0.425 & 0.911 & 0.425 & 0.934 \\
$[0.45, 0.50)$ & 0.475 & 0.900 & 0.475 & 0.930 \\
$[0.50, 0.55)$ & 0.525 & 0.885 & 0.525 & 0.915 \\
$[0.55, 0.60)$ & 0.575 & 0.870 & 0.575 & 0.897 \\
$[0.60, 0.65)$ & 0.625 & 0.838 & 0.626 & 0.879 \\
$[0.65, 0.70)$ & 0.675 & 0.806 & 0.676 & 0.882 \\
$[0.70, 0.75)$ & 0.725 & 0.766 & 0.726 & 0.886 \\
$[0.75, 0.80)$ & 0.774 & 0.694 & 0.775 & 0.868 \\
$[0.80, 0.85)$ & 0.825 & 0.592 & 0.825 & 0.827 \\
$[0.85, 0.90)$ & 0.876 & 0.448 & 0.875 & 0.797 \\
$[0.90, 0.95)$ & 0.928 & 0.291 & 0.928 & 0.777 \\
$[0.95, 1.00)$ & 0.979 & 0.282 & 0.978 & 0.790 \\
\bottomrule
\end{tabular}
}
\end{table*}

\begin{table*}[h]
\centering
\caption{Dice results of MedSAM (Union Box) and PathSegmentor on intricate objects with decreasing instance ratio. The column of Instance Ratio Interval indicates the interval partitioning of the instance ratio. Avg. Instance Ratio and Avg. Dice represent the weighted average values within each interval, where the weights are determined by Gaussian Kernel Density Estimation based on the density of data points.}
\label{tab_supp:intricate_performance_tiny}
\resizebox{\textwidth}{!}{%
\begin{tabular}{lcccc}
\toprule
\multirow{2}{*}[-0.8ex]{\makecell{\textbf{Instance Ratio}\\ \textbf{Interval}}} & \multicolumn{2}{c}{\textbf{MedSAM (Union Box)}} & \multicolumn{2}{c}{\textbf{PathSegmentor}} \\
\cmidrule(lr){2-3}\cmidrule(lr){4-5}
& Avg. Instance Ratio & Avg. Dice & Avg. Instance Ratio & Avg. Dice \\
\midrule
$[0.95, 1.00)$ & 0.993 & 0.976 & 0.993 & 0.996 \\
$[0.90, 0.95)$ & 0.929 & 0.949 & 0.929 & 0.977 \\
$[0.85, 0.90)$ & 0.877 & 0.925 & 0.877 & 0.968 \\
$[0.80, 0.85)$ & 0.826 & 0.905 & 0.827 & 0.967 \\
$[0.75, 0.80)$ & 0.776 & 0.895 & 0.776 & 0.962 \\
$[0.70, 0.75)$ & 0.725 & 0.885 & 0.725 & 0.962 \\
$[0.65, 0.70)$ & 0.676 & 0.884 & 0.675 & 0.956 \\
$[0.60, 0.65)$ & 0.625 & 0.901 & 0.624 & 0.959 \\
$[0.55, 0.60)$ & 0.573 & 0.901 & 0.574 & 0.956 \\
$[0.50, 0.55)$ & 0.523 & 0.899 & 0.524 & 0.952 \\
$[0.45, 0.50)$ & 0.471 & 0.920 & 0.471 & 0.964 \\
$[0.40, 0.45)$ & 0.425 & 0.900 & 0.425 & 0.954 \\
$[0.35, 0.40)$ & 0.376 & 0.864 & 0.375 & 0.947 \\
$[0.30, 0.35)$ & 0.325 & 0.839 & 0.325 & 0.941 \\
$[0.25, 0.30)$ & 0.273 & 0.835 & 0.273 & 0.938 \\
$[0.20, 0.25)$ & 0.225 & 0.817 & 0.224 & 0.928 \\
$[0.15, 0.20)$ & 0.174 & 0.795 & 0.173 & 0.918 \\
$[0.10, 0.15)$ & 0.123 & 0.763 & 0.123 & 0.898 \\
$[0.05, 0.10)$ & 0.072 & 0.680 & 0.072 & 0.839 \\
$[0.00, 0.05)$ & 0.004 & 0.159 & 0.007 & 0.531 \\
\bottomrule
\end{tabular}
}
\end{table*}

\begin{table*}[h]
\centering
\caption{Dice results of MedSAM (Union Box) and PathSegmentor on intricate objects with increasing instance count. The column of Instance Count Interval indicates the interval partitioning of the instance count. Avg. Instance Count and Avg. Dice represent the weighted average values within each interval, where the weights are determined by Gaussian Kernel Density Estimation based on the density of data points.}
\label{tab_supp:intricate_performance_dense}
\resizebox{\textwidth}{!}{%
\begin{tabular}{lcccc}
\toprule
\multirow{2}{*}[-0.8ex]{\makecell{\textbf{Instance Count}\\\textbf{Interval}}} & \multicolumn{2}{c}{\textbf{MedSAM (Union Box)}} & \multicolumn{2}{c}{\textbf{PathSegmentor}} \\
\cmidrule(lr){2-3}\cmidrule(lr){4-5}
& Avg. Instance Count & Avg. Dice & Avg. Instance Count & Avg. Dice \\
\midrule
$[2, 4)$ & 2.343 & 0.492 & 2.307 & 0.813 \\
$[4, 6)$ & 4.394 & 0.218 & 4.340 & 0.719 \\
$[6, 8)$ & 6.439 & 0.116 & 6.400 & 0.680 \\
$[8, 10)$ & 8.455 & 0.077 & 8.437 & 0.660 \\
$[10, 12)$ & 10.450 & 0.065 & 10.459 & 0.656 \\
$[12, 14)$ & 12.449 & 0.065 & 12.445 & 0.674 \\
\bottomrule
\end{tabular}
}
\end{table*}

\begin{table*}[h]
\centering
\caption{Prompt Efficiency of PathSegmentor and MedSAM. The column of \#Prompts indicates the average number of prompts used per mask, while Dice represents the average Dice score for that group.}
\label{tab_supp:intricate_performance_vs_spatial_prompt}
\resizebox{\textwidth}{!}{%
\begin{tabular}{lcccccccc}
\toprule
\multirow{2}{*}[-0.8ex]{\makecell{}} & \multicolumn{2}{c}{\textbf{MedSAM (Union Box)}} & \multicolumn{2}{c}{\textbf{MedSAM (Instance Box)}} & \multicolumn{2}{c}{\textbf{Instance Box}} & \multicolumn{2}{c}{\textbf{PathSegmentor}} \\
\cmidrule(lr){2-3}\cmidrule(lr){4-5}\cmidrule(lr){6-7}\cmidrule(lr){8-9}
& \#Prompts & Dice & \#Prompts & Dice & \#Prompts & Dice & \#Prompts & Dice\\
\midrule
Tissue & 1.0 & 0.678 & 11.6 & 0.793 & 11.6 & 0.788 & 1.0 & 0.747 \\
Cell & 1.0 & 0.121 & 28.2 & 0.662 & 28.2 & 0.767 & 1.0 & 0.382 \\
Nuclei & 1.0 & 0.284 & 16.0 & 0.791 & 16.0 & 0.841 & 1.0 & 0.614 \\
Overall & 1.0 & 0.526 & 14.7 & 0.774 & 14.7 & 0.795 & 1.0 & 0.671 \\
\bottomrule
\end{tabular}
}
\end{table*}
\begin{table*}[t]
\centering
\caption{Dice scores for segmentation results across different anatomical regions (AR) on the internal datasets of PathSeg. \textbf{Bold} indicates the best result and \underline{underline} indicates the second best. 95\% confidence interval (CI) is included in square brackets. P-values are computed using one-sided, paired Student’s $t$-tests between PathSegmentor and the second-best models.}
\resizebox{\textwidth}{!}{%
\begin{tabular}{lccccccccc}
\toprule
\multirow{2}{*}[-2.ex]{\textbf{AR}} & \multicolumn{3}{c}{\textbf{Specialized Model}} & \multicolumn{4}{c}{\textbf{Segmentation Foundation Model}} & \multirow{2}{*}[-2.ex]{\textbf{\textit{P}-value}}\\
            \cmidrule(lr){2-4} \cmidrule(lr){5-8}
     & \textbf{DeepLabV3+} & \textbf{SAM-Path} & \textbf{nnU-Net} & \makecell{\textbf{MedSAM} \\ \textbf{(Union Box)}} & \makecell{\textbf{SAM-Med2D} \\ \textbf{(Union Box)}} & \textbf{BiomedParse} & \textbf{PathSegmentor} \\
\midrule
Lung & \makecell{0.660 \\ \scriptsize{[0.644, 0.675]}} & \makecell{0.727 \\ \scriptsize{[0.712, 0.740]}} & \makecell{\underline{0.811} \\ \scriptsize{[0.798, 0.825]}} & \makecell{0.399 \\ \scriptsize{[0.385, 0.412]}} & \makecell{0.373 \\ \scriptsize{[0.359, 0.387]}} & \makecell{0.068 \\ \scriptsize{[0.060, 0.077]}} & \makecell{\textbf{0.813} \\ \scriptsize{[0.801, 0.824]}} & $\leq 0.01$ \\
Prostate & \makecell{0.603 \\ \scriptsize{[0.598, 0.609]}} & \makecell{0.720 \\ \scriptsize{[0.716, 0.724]}} & \makecell{0.651 \\ \scriptsize{[0.646, 0.657]}} & \makecell{\underline{0.781} \\ \scriptsize{[0.778, 0.784]}} & \makecell{0.613 \\ \scriptsize{[0.610, 0.616]}} & \makecell{0.246 \\ \scriptsize{[0.243, 0.249]}} & \makecell{\textbf{0.805} \\ \scriptsize{[0.801, 0.809]}} & $\leq 0.01$ \\
Colon & \makecell{0.545 \\ \scriptsize{[0.539, 0.552]}} & \makecell{0.438 \\ \scriptsize{[0.430, 0.446]}} & \makecell{\underline{0.660} \\ \scriptsize{[0.653, 0.666]}} & \makecell{0.443 \\ \scriptsize{[0.435, 0.451]}} & \makecell{0.371 \\ \scriptsize{[0.364, 0.378]}} & \makecell{0.407 \\ \scriptsize{[0.400, 0.414]}} & \makecell{\textbf{0.727} \\ \scriptsize{[0.722, 0.733]}} & $\leq 0.01$ \\
\bottomrule
\end{tabular}
}
\label{tab_supp:performance_across_organs}
\end{table*}

\begin{table*}[t]
\centering
\caption{Dice scores for segmentation results across different histological structures (HS) on the internal datasets of PathSeg. \textbf{Bold} indicates the best result and \underline{underline} indicates the second best. 95\% confidence interval (CI) is included in square brackets. P-values are computed using one-sided, paired Student’s $t$-tests between PathSegmentor and the second-best models.
}
\resizebox{\textwidth}{!}{%
\begin{tabular}{lcccccccc}
\toprule
\multirow{2}{*}[-2.ex]{\textbf{HS}} & \multicolumn{3}{c}{\textbf{Specialized Model}} & \multicolumn{4}{c}{\textbf{Segmentation Foundation Model}} & \multirow{2}{*}[-2.ex]{\textbf{\textit{P}-value}}\\
            \cmidrule(lr){2-4} \cmidrule(lr){5-8}
     & \textbf{DeepLabV3+} & \textbf{SAM-Path} & \textbf{nnU-Net} & \makecell{\textbf{MedSAM} \\ \textbf{(Union Box)}} & \makecell{\textbf{SAM-Med2D} \\ \textbf{(Union Box)}} & \textbf{BiomedParse} & \textbf{PathSegmentor} \\
\midrule
Tissue & \makecell{0.547 \\ \scriptsize{[0.543, 0.551]}} & \makecell{0.609 \\ \scriptsize{[0.604, 0.613]}} & \makecell{0.570 \\ \scriptsize{[0.566, 0.575]}} & \makecell{\underline{0.678} \\ \scriptsize{[0.675, 0.682]}} & \makecell{0.551 \\ \scriptsize{[0.549, 0.555]}} & \makecell{0.240 \\ \scriptsize{[0.237, 0.242]}} & \makecell{\textbf{0.747} \\ \scriptsize{[0.744, 0.751]}} & $\leq 0.01$ \\
Nuclei & \makecell{0.452 \\ \scriptsize{[0.445, 0.458]}} & \makecell{0.322 \\ \scriptsize{[0.316, 0.329]}} & \makecell{\underline{0.547} \\ \scriptsize{[0.541, 0.554]}} & \makecell{0.284 \\ \scriptsize{[0.278, 0.289]}} & \makecell{0.252 \\ \scriptsize{[0.247, 0.257]}} & \makecell{0.395 \\ \scriptsize{[0.388, 0.402]}} & \makecell{\textbf{0.614} \\ \scriptsize{[0.607, 0.620]}} & $\leq 0.01$ \\
Cell & \makecell{0.067 \\ \scriptsize{[0.063, 0.071]}} & \makecell{0.016 \\ \scriptsize{[0.014, 0.019]}} & \makecell{0.112 \\ \scriptsize{[0.106, 0.117]}} & \makecell{\underline{0.121} \\ \scriptsize{[0.116, 0.127]}} & \makecell{0.100 \\ \scriptsize{[0.096, 0.105]}} & \makecell{0.048 \\ \scriptsize{[0.045, 0.050]}} & \makecell{\textbf{0.382} \\ \scriptsize{[0.375, 0.388]}} & $\leq 0.01$ \\

\bottomrule
\end{tabular}
}
\label{tab_supp:performance_across_scales}
\end{table*}
\clearpage
{
\setlength\LTleft{-1.3in}
\setlength\LTright{-1.3in plus 1 fill}
\captionsetup[longtable]{width=10in, justification=raggedright, margin={-2.cm,-2.1cm}}
\newcommand{\mysmall}{\fontsize{6pt}{7pt}\selectfont}
\newcommand{\mytiny}{\fontsize{5pt}{6pt}\selectfont}
\mysmall
\begin{longtable}{
    p{4.cm}                                     
    >{\centering\arraybackslash}p{1.5cm}         
    >{\centering\arraybackslash}p{1.5cm}         
    >{\centering\arraybackslash}p{1.5cm}         
    >{\centering\arraybackslash}p{1.6cm}
    >{\centering\arraybackslash}p{1.7cm}
    >{\centering\arraybackslash}p{1.6cm}
    >{\centering\arraybackslash}p{1.7cm}
    >{\centering\arraybackslash}p{1.0cm}
}
    \caption{Dice scores for segmentation results across different object types (OT) on the internal datasets of PathSeg. 
    \textbf{Bold} indicates the best result and \underline{underline} indicates the second best. 95\% confidence interval (CI) is included in square brackets. P-values are computed using one-sided, paired Student’s $t$-tests between PathSegmentor and the second-best models.}\\
    
     \toprule
     \multirow{2}{*}[-2.ex]{\textbf{OT}} & \multicolumn{3}{c}{\textbf{Specialized Model}} & \multicolumn{4}{c}{\textbf{Segmentation Foundation Model}}  & \multirow{2}{*}[-2.ex]{\textbf{\textit{P}-value}}\\
                 \cmidrule(lr){2-4} \cmidrule(lr){5-8}
          & \textbf{DeepLabV3+} & \textbf{SAM-Path} & \textbf{nnU-Net} & \makecell{\textbf{MedSAM} \\ \textbf{(Union Box)}} & \makecell{\textbf{SAM-Med2D} \\ \textbf{(Union Box)}} & \textbf{BiomedParse} & \textbf{PathSegmentor} \\
     \midrule
     \endfirsthead
    \multicolumn{8}{c}{{\bfseries \tablename\ \thetable{} -- continued from previous page}} \\
     \toprule
     \multirow{2}{*}[-2.ex]{\textbf{OT}} & \multicolumn{3}{c}{\textbf{Specialized Model}} & \multicolumn{4}{c}{\textbf{Segmentation Foundation Model}} \\
                 \cmidrule(lr){2-4} \cmidrule(lr){5-8}
          & \textbf{DeepLabV3+} & \textbf{SAM-Path} & \textbf{nnU-Net} & \makecell{\textbf{MedSAM} \\ \textbf{(Union Box)}} & \makecell{\textbf{SAM-Med2D} \\ \textbf{(Union Box)}} & \textbf{BiomedParse} & \textbf{PathSegmentor} \\
     \midrule
     \endhead
     \midrule
     \multicolumn{8}{r}{Continued on next page} \\
     \endfoot
     \bottomrule
     \endlastfoot
Gland & \makecell{0.749 \\ \mytiny{[0.731, 0.765]}} & \makecell{0.818 \\ \mytiny{[0.805, 0.829]}} & \makecell{\underline{0.897} \\ \mytiny{[0.883, 0.909]}} & \makecell{0.780 \\ \mytiny{[0.763, 0.796]}} & \makecell{0.596 \\ \mytiny{[0.577, 0.613]}} & \makecell{0.619 \\ \mytiny{[0.592, 0.646]}} & \makecell{\textbf{0.934} \\ \mytiny{[0.927, 0.940]}} & $\leq 0.01$ \\
Tumor epithelial tissue & \makecell{0.803 \\ \mytiny{[0.792, 0.815]}} & \makecell{0.839 \\ \mytiny{[0.827, 0.849]}} & \makecell{\textbf{0.918} \\ \mytiny{[0.908, 0.926]}} & \makecell{0.336 \\ \mytiny{[0.318, 0.354]}} & \makecell{0.239 \\ \mytiny{[0.227, 0.253]}} & \makecell{0.007 \\ \mytiny{[0.006, 0.009]}} & \makecell{\underline{0.884} \\ \mytiny{[0.873, 0.893]}} & - \\
Prostate tumor & \makecell{0.572 \\ \mytiny{[0.563, 0.582]}} & \makecell{0.772 \\ \mytiny{[0.765, 0.780]}} & \makecell{0.745 \\ \mytiny{[0.736, 0.753]}} & \makecell{\underline{0.841} \\ \mytiny{[0.836, 0.845]}} & \makecell{0.657 \\ \mytiny{[0.652, 0.662]}} & \makecell{0.307 \\ \mytiny{[0.303, 0.310]}} & \makecell{\textbf{0.866} \\ \mytiny{[0.860, 0.871]}} & $\leq 0.01$ \\
Metastatic colon cancer in a liver & \makecell{0.663 \\ \mytiny{[0.582, 0.736]}} & \makecell{0.757 \\ \mytiny{[0.704, 0.806]}} & \makecell{\underline{0.822} \\ \mytiny{[0.755, 0.881]}} & \makecell{0.611 \\ \mytiny{[0.518, 0.692]}} & \makecell{0.509 \\ \mytiny{[0.388, 0.609]}} & \makecell{0.679 \\ \mytiny{[0.615, 0.734]}} & \makecell{\textbf{0.827} \\ \mytiny{[0.791, 0.861]}} & $0.06$ \\
Prostate cancer & \makecell{0.608 \\ \mytiny{[0.597, 0.618]}} & \makecell{0.682 \\ \mytiny{[0.674, 0.691]}} & \makecell{0.622 \\ \mytiny{[0.612, 0.632]}} & \makecell{\underline{0.806} \\ \mytiny{[0.801, 0.811]}} & \makecell{0.700 \\ \mytiny{[0.694, 0.706]}} & \makecell{0.241 \\ \mytiny{[0.235, 0.246]}} & \makecell{\textbf{0.821} \\ \mytiny{[0.813, 0.828]}} & $\leq 0.01$ \\
\end{longtable}
\normalsize

\setlength\LTleft{0in}
\setlength\LTright{0in}
\setlength{\LTcapwidth}{0in}
}
\clearpage
{
\setlength\LTleft{-1.3in}
\setlength\LTright{-1.3in plus 1 fill}
\captionsetup[longtable]{width=10in, justification=raggedright, margin={-2.cm,-2.1cm}}
\newcommand{\mysmall}{\fontsize{6pt}{7pt}\selectfont}
\newcommand{\mytiny}{\fontsize{5pt}{6pt}\selectfont}
\mysmall
\begin{longtable}{
    p{5cm}                                     
    >{\centering\arraybackslash}p{1.5cm}         
    >{\centering\arraybackslash}p{1.5cm}         
    >{\centering\arraybackslash}p{1.5cm}         
    >{\centering\arraybackslash}p{1.6cm}
    >{\centering\arraybackslash}p{1.7cm}
    >{\centering\arraybackslash}p{1.6cm}
    >{\centering\arraybackslash}p{1.7cm}
}
    \caption{Dice scores for segmentation results of each category on the internal datasets of PathSeg. Overall results are reported by averaging the Dice scores of all samples on the internal datasets.
    \textbf{Bold} indicates the best result and \underline{underline} indicates the second best. 95\% confidence interval (CI) is included in square brackets.}\\
    
     \toprule
     \multirow{2}{*}[-2.ex]{\textbf{Category}} & \multicolumn{3}{c}{\textbf{Specialized Model}} & \multicolumn{4}{c}{\textbf{Segmentation Foundation Model}} \\
                 \cmidrule(lr){2-4} \cmidrule(lr){5-8}
          & \textbf{DeepLabV3+} & \textbf{SAM-Path} & \textbf{nnU-Net} & \makecell{\textbf{MedSAM} \\ \textbf{(Union Box)}} & \makecell{\textbf{SAM-Med2D} \\ \textbf{(Union Box)}} & \textbf{BiomedParse} & \textbf{PathSegmentor} \\
     \midrule
     \endfirsthead
    \multicolumn{8}{c}{{\bfseries \tablename\ \thetable{} -- continued from previous page}} \\
     \toprule
     \multirow{2}{*}[-2.ex]{\textbf{Object Type}} & \multicolumn{3}{c}{\textbf{Specialized Model}} & \multicolumn{4}{c}{\textbf{Segmentation Foundation Model}} \\
                 \cmidrule(lr){2-4} \cmidrule(lr){5-8}
          & \textbf{DeepLabV3+} & \textbf{SAM-Path} & \textbf{nnU-Net} & \makecell{\textbf{MedSAM} \\ \textbf{(Union Box)}} & \makecell{\textbf{SAM-Med2D} \\ \textbf{(Union Box)}} & \textbf{BiomedParse} & \textbf{PathSegmentor} \\
     \midrule
     \endhead
     \midrule
     \multicolumn{8}{r}{Continued on next page} \\
     \endfoot
     \bottomrule
     \endlastfoot
Cervix-Nuclei-Epithelial & \makecell{0.559 \\ \mytiny{[0.159, 0.927]}} & \makecell{0.000 \\ \mytiny{[0.000, 0.000]}} & \makecell{0.071 \\ \mytiny{[0.000, 0.213]}} & \makecell{0.446 \\ \mytiny{[0.256, 0.796]}} & \makecell{0.446 \\ \mytiny{[0.161, 0.953]}} & \makecell{\underline{0.627} \\ \mytiny{[0.000, 0.948]}} & \makecell{\textbf{0.953} \\ \mytiny{[0.928, 0.967]}} \\
Colon-Tissue-Gland & \makecell{0.749 \\ \mytiny{[0.731, 0.765]}} & \makecell{0.818 \\ \mytiny{[0.805, 0.829]}} & \makecell{\underline{0.897} \\ \mytiny{[0.883, 0.909]}} & \makecell{0.780 \\ \mytiny{[0.763, 0.796]}} & \makecell{0.596 \\ \mytiny{[0.577, 0.613]}} & \makecell{0.619 \\ \mytiny{[0.592, 0.646]}} & \makecell{\textbf{0.934} \\ \mytiny{[0.927, 0.940]}} \\
Lung-Tissue-Tumor epithelial tissue & \makecell{0.803 \\ \mytiny{[0.792, 0.815]}} & \makecell{0.839 \\ \mytiny{[0.827, 0.849]}} & \makecell{\textbf{0.918} \\ \mytiny{[0.908, 0.926]}} & \makecell{0.336 \\ \mytiny{[0.318, 0.354]}} & \makecell{0.239 \\ \mytiny{[0.227, 0.253]}} & \makecell{0.007 \\ \mytiny{[0.006, 0.009]}} & \makecell{\underline{0.884} \\ \mytiny{[0.873, 0.893]}} \\
Liver-Nuclei-Epithelial & \makecell{0.818 \\ \mytiny{[0.720, 0.882]}} & \makecell{0.000 \\ \mytiny{[0.000, 0.000]}} & \makecell{0.628 \\ \mytiny{[0.405, 0.794]}} & \makecell{0.207 \\ \mytiny{[0.153, 0.262]}} & \makecell{0.173 \\ \mytiny{[0.116, 0.229]}} & \makecell{\underline{0.871} \\ \mytiny{[0.823, 0.903]}} & \makecell{\textbf{0.875} \\ \mytiny{[0.828, 0.906]}} \\
Testis-Nuclei-Epithelial & \makecell{0.856 \\ \mytiny{[0.832, 0.885]}} & \makecell{0.000 \\ \mytiny{[0.000, 0.000]}} & \makecell{0.787 \\ \mytiny{[0.627, 0.887]}} & \makecell{0.418 \\ \mytiny{[0.285, 0.542]}} & \makecell{0.356 \\ \mytiny{[0.238, 0.460]}} & \makecell{\textbf{0.879} \\ \mytiny{[0.850, 0.909]}} & \makecell{\underline{0.868} \\ \mytiny{[0.822, 0.910]}} \\
\end{longtable}
\normalsize

\setlength\LTleft{0in}
\setlength\LTright{0in}
\setlength{\LTcapwidth}{0in}
}
\begin{figure}[H]
    \centering
    \includegraphics[width=\linewidth]{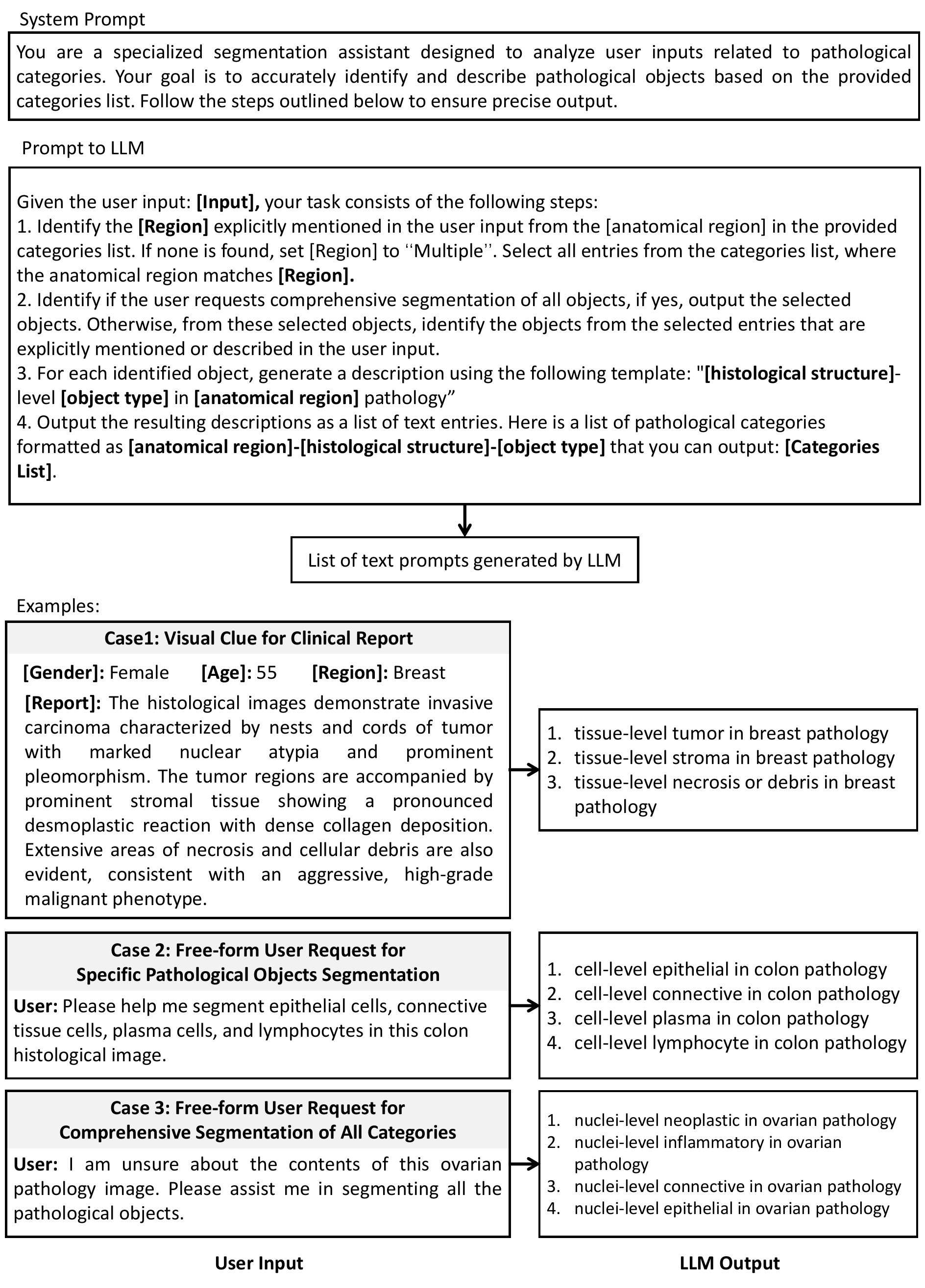}
    \caption{Prompts of Large Language Model (LLM) that are used to generate lists of text inputs for PathSegmentor.
    }
    \label{fig_supp:application_prompt}
\end{figure}

\begin{table*}[h]
\centering
\caption{The correspondence between external and internal categories, along with the performance of specialized models in external evaluation.}
\resizebox{\textwidth}{!}{%
\begin{tabular}{lccccccc}
\toprule
\textbf{External Dataset} & \textbf{External Category} & \textbf{Training Dataset} & \textbf{Internal Category} & \textbf{DeepLabV3+} & \textbf{SAM-Path} & \textbf{nnU-Net} \\
\midrule
\multirow{11}{*}{\raisebox{-11.\normalbaselineskip}{CoNSeP}}
    & Colon-Cell-Malignant epithelial & WSSS4LUAD & Tumor epithelial tissue & \makecell{\textbf{0.002} \\ \scriptsize{[0.001, 0.003]}} & \makecell{\textbf{0.004} \\ \scriptsize{[0.001, 0.006]}} & \makecell{\textbf{0.003} \\ \scriptsize{[0.000, 0.005]}} \\
\cmidrule{2-7}
    & \multirow{2}{*}{\raisebox{-0.5\normalbaselineskip}{Colon-Cell-Endothelial}} & SegPath & Endothelial & \makecell{0.000 \\ \scriptsize{[0.000, 0.000]}} & \makecell{0.000 \\ \scriptsize{[0.000, 0.000]}} & \makecell{0.000 \\ \scriptsize{[0.000, 0.000]}} \\
    & & TNBC & Endothelial & \makecell{\textbf{0.005} \\ \scriptsize{[-0.002, 0.010]}} & \makecell{\textbf{0.000} \\ \scriptsize{[0.000, 0.000]}} & \makecell{\textbf{0.006} \\ \scriptsize{[-0.004, 0.013]}} \\
\cmidrule{2-7}
    & \multirow{2}{*}{\raisebox{-0.5\normalbaselineskip}{Colon-Cell-Fibroblast}} & NuCLS & Fibroblast & \makecell{0.011 \\ \scriptsize{[0.002, 0.018]}} & \makecell{\textbf{0.033} \\ \scriptsize{[0.015, 0.049]}} & \makecell{\textbf{0.141} \\ \scriptsize{[0.093, 0.187]}} \\
    & & TNBC & Fibroblast & \makecell{\textbf{0.014} \\ \scriptsize{[0.007, 0.020]}} & \makecell{0.000 \\ \scriptsize{[0.000, 0.000]}} & \makecell{0.093 \\ \scriptsize{[0.054, 0.128]}} \\
\cmidrule{2-7}
    & \multirow{4}{*}{\raisebox{-2.5\normalbaselineskip}{Colon-Cell-Healthy epithelial}} & MoNuSAC2020 & Epithelial & \makecell{\textbf{0.002} \\ \scriptsize{[-0.002, 0.003]}} & \makecell{0.109 \\ \scriptsize{[-0.041, 0.215]}} & \makecell{\textbf{0.159} \\ \scriptsize{[0.045, 0.272]}} \\
    & & PanNuke & Epithelial & \makecell{0.000 \\ \scriptsize{[0.000, 0.000]}} & \makecell{0.000 \\ \scriptsize{[0.000, 0.000]}} & \makecell{0.000 \\ \scriptsize{[0.000, 0.000]}} \\
\bottomrule
\end{tabular}
}
\label{tab_supp:external_category}
\end{table*}

\addvspace{-10pt}
\begin{table*}[t]
\centering
\caption{Dice scores for segmentation results on the external datasets of PathSeg. For the CoNSeP dataset, we present results for the overall dataset as well as for each category separately. For the other datasets, since they include only one nuclei category, we provide results at the dataset level. \textbf{Bold} indicates the best result and \underline{underline} indicates the second best. 95\% confidence interval (CI) is included in square brackets. P-values are computed using one-sided, paired Student’s $t$-tests between PathSegmentor and the second-best models.}
\resizebox{\textwidth}{!}{%
\begin{tabular}{lcccccccc}
\toprule
\multirow{2}{*}[-2.ex]{\textbf{Dataset/Category}} & \multicolumn{3}{c}{\textbf{Specialized Model}} & \multicolumn{4}{c}{\textbf{Segmentation Foundation Model}} & \multirow{2}{*}[-2.ex]{\textbf{\textit{P}-value}}\\
        \cmidrule(lr){2-4} \cmidrule(lr){5-8}
     & \textbf{DeepLabV3+} & \textbf{SAM-Path} & \textbf{nnU-Net} & \makecell{\textbf{MedSAM} \\ \textbf{(Union Box)}} & \makecell{\textbf{SAM-Med2D} \\ \textbf{(Union Box)}} & \textbf{BiomedParse} & \textbf{PathSegmentor} \\
\midrule
CPM17 
& \makecell{0.052 \\ \scriptsize{[0.036, 0.067]}}
& \makecell{0.274 \\ \scriptsize{[0.242, 0.306]}}
& \makecell{0.111 \\ \scriptsize{[0.095, 0.129]}}
& \makecell{0.329 \\ \scriptsize{[0.299, 0.355]}}
& \makecell{0.316 \\ \scriptsize{[0.292, 0.342]}}
& \makecell{\underline{0.553} \\ \scriptsize{[0.485, 0.619]}}
& \makecell{\textbf{0.705} \\ \scriptsize{[0.670, 0.738]}}
& $\leq 0.01$ \\
CPM15 
& \makecell{0.056 \\ \scriptsize{[0.030, 0.086]}}
& \makecell{0.211 \\ \scriptsize{[0.157, 0.273]}}
& \makecell{0.132 \\ \scriptsize{[0.092, 0.170]}}
& \makecell{0.274 \\ \scriptsize{[0.234, 0.308]}}
& \makecell{\underline{0.278} \\ \scriptsize{[0.244, 0.312]}}
& \makecell{0.236 \\ \scriptsize{[0.087, 0.396]}}
& \makecell{\textbf{0.485} \\ \scriptsize{[0.365, 0.602]}}
& $\leq 0.01$ \\
Kumar 
& \makecell{0.067 \\ \scriptsize{[0.045, 0.094]}}
& \makecell{0.170 \\ \scriptsize{[0.133, 0.206]}}
& \makecell{0.083 \\ \scriptsize{[0.059, 0.110]}}
& \makecell{\underline{0.393} \\ \scriptsize{[0.351, 0.435]}}
& \makecell{0.368 \\ \scriptsize{[0.326, 0.408]}}
& \makecell{0.181 \\ \scriptsize{[0.093, 0.269]}}
& \makecell{\textbf{0.427} \\ \scriptsize{[0.387, 0.465]}}
& $\leq 0.01$ \\
Lizard 
& \makecell{0.010 \\ \scriptsize{[0.007, 0.015]}}
& \makecell{0.044 \\ \scriptsize{[0.037, 0.051]}}
& \makecell{0.017 \\ \scriptsize{[0.013, 0.022]}}
& \makecell{\underline{0.299} \\ \scriptsize{[0.286, 0.310]}}
& \makecell{0.282 \\ \scriptsize{[0.271, 0.293]}}
& \makecell{0.232 \\ \scriptsize{[0.212, 0.254]}}
& \makecell{\textbf{0.319} \\ \scriptsize{[0.304, 0.333]}}
& $\leq 0.01$ \\
\midrule
CoNSeP & \makecell{0.005 \\ \scriptsize{[0.003, 0.008]}} & \makecell{0.030 \\ \scriptsize{[0.019, 0.044]}} & \makecell{0.054 \\ \scriptsize{[0.037, 0.075]}} & \makecell{0.170 \\ \scriptsize{[0.136, 0.206]}} & \makecell{0.148 \\ \scriptsize{[0.120, 0.177]}} & \makecell{\underline{0.217} \\ \scriptsize{[0.170, 0.263]}} & \makecell{\textbf{0.291} \\ \scriptsize{[0.244, 0.335]}} & $\leq 0.01$\\
Malignant epithelial & \makecell{0.002 \\ \scriptsize{[0.001, 0.003]}} & \makecell{0.004 \\ \scriptsize{[0.002, 0.007]}} & \makecell{0.003 \\ \scriptsize{[0.001, 0.006]}} & \makecell{0.462 \\ \scriptsize{[0.409, 0.524]}} & \makecell{0.406 \\ \scriptsize{[0.366, 0.452]}} & \makecell{\underline{0.615} \\ \scriptsize{[0.528, 0.689]}} & \makecell{\textbf{0.624} \\ \scriptsize{[0.526, 0.702]}} & $\leq 0.01$ \\
Healthy epithelial & \makecell{0.002 \\ \scriptsize{[0.000, 0.005]}} & \makecell{0.270 \\ \scriptsize{[0.217, 0.316]}} & \makecell{0.159 \\ \scriptsize{[0.046, 0.272]}} & \makecell{0.265 \\ \scriptsize{[0.218, 0.335]}} & \makecell{0.221 \\ \scriptsize{[0.185, 0.252]}} & \makecell{\underline{0.383} \\ \scriptsize{[0.297, 0.463]}} & \makecell{\textbf{0.596} \\ \scriptsize{[0.452, 0.715]}} & $\leq 0.01$ \\
Inflammatory & \makecell{0.000 \\ \scriptsize{[0.000, 0.000]}} & \makecell{0.000 \\ \scriptsize{[0.000, 0.000]}} & \makecell{0.011 \\ \scriptsize{[0.002, 0.025]}} & \makecell{0.127 \\ \scriptsize{[0.067, 0.204]}} & \makecell{0.107 \\ \scriptsize{[0.060, 0.166]}} & \makecell{\underline{0.197} \\ \scriptsize{[0.131, 0.265]}} & \makecell{\textbf{0.301} \\ \scriptsize{[0.217, 0.382]}} & $\leq 0.01$ \\
Muscle & \makecell{0.000 \\ \scriptsize{[0.000, 0.000]}} & \makecell{0.089 \\ \scriptsize{[0.060, 0.119]}} & \makecell{0.000 \\ \scriptsize{[0.000, 0.000]}} & \makecell{0.111 \\ \scriptsize{[0.093, 0.130]}} & \makecell{0.114 \\ \scriptsize{[0.094, 0.131]}} & \makecell{\underline{0.267} \\ \scriptsize{[0.113, 0.425]}} & \makecell{\textbf{0.301} \\ \scriptsize{[0.178, 0.432]}} & $\leq 0.01$ \\
Fibroblast & \makecell{0.014 \\ \scriptsize{[0.008, 0.021]}} & \makecell{0.033 \\ \scriptsize{[0.017, 0.052]}} & \makecell{\underline{0.141} \\ \scriptsize{[0.095, 0.183]}} & \makecell{0.093 \\ \scriptsize{[0.068, 0.120]}} & \makecell{0.090 \\ \scriptsize{[0.066, 0.116]}} & \makecell{0.065 \\ \scriptsize{[0.036, 0.098]}} & \makecell{\textbf{0.148} \\ \scriptsize{[0.106, 0.188]}} & $\leq 0.01$ \\
Endothelial & \makecell{0.005 \\ \scriptsize{[0.000, 0.012]}} & \makecell{0.000 \\ \scriptsize{[0.000, 0.000]}} & \makecell{0.006 \\ \scriptsize{[0.000, 0.015]}} & \makecell{\textbf{0.121} \\ \scriptsize{[0.035, 0.226]}} & \makecell{\underline{0.070} \\ \scriptsize{[0.030, 0.111]}} & \makecell{0.032 \\ \scriptsize{[0.000, 0.090]}} & \makecell{0.034 \\ \scriptsize{[0.004, 0.085]}} & - \\

\bottomrule
\end{tabular}
}
\label{tab_supp:performance_external}
\end{table*}

\begin{table*}[t]
\centering
\caption{Mean and standard deviation of number of instances (Num Ins.) and maximum inter-instance distance (Max Dis.) for each category on the external dataset.}
\label{tab_supp:external_instance_stats}
\begin{tabular}{lrrrr}
\toprule
\multirow{2}{*}{\textbf{Category}} & \multicolumn{2}{c}{\textbf{Num Ins.}} & \multicolumn{2}{c}{\textbf{Max Dis.}}\\
\cmidrule{2-3}
\cmidrule{4-5}
& \textbf{Mean} & \textbf{Std} & \textbf{Mean} & \textbf{Std} \\
\midrule
Endothelial & 8.09 & 13.71 & 425.42 & 425.38 \\
Malignant epithelial & 176.39 & 106.50 & 1112.17 & 363.26 \\
Fibroblast & 151.00 & 137.01 & 1181.19 & 285.81 \\
Healthy epithelial & 155.40 & 31.50 & 1319.89 & 42.49 \\
Inflammatory & 107.90 & 204.58 & 989.85 & 445.49 \\
Muscle & 106.36 & 79.53 & 1140.96 & 233.92 \\
\bottomrule
\end{tabular}
\end{table*}

\begin{table*}[h]
\centering
\caption{Results (mean $\pm$ std) of the standard model and object-based model for breast cancer diagnosis on the TCGA-BRCA dataset. }
\label{tab_supp:diagnosis_result}
\begin{tabular}{lccc}
\toprule
Method & Macro AUC & Macro ACC & Macro F1 \\
\midrule
Standard model &~~0.936$\pm$0.017~~ & ~~0.825$\pm$0.027~~ & ~~0.835$\pm$0.024~~\\
Object-based model & 0.953$\pm$0.014 & 0.857$\pm$0.024 & 0.868$\pm$0.022\\
\bottomrule
\end{tabular}
\end{table*}

\begin{table*}[h]
\centering
\caption{Lists of pathological objects utilized in the three explainability analysis experiments.}
\label{tab_supp:xai_object}
\begin{tabular}{
    m{3cm}                                     
    >{\centering\arraybackslash}m{1cm}         
    >{\arraybackslash}m{8cm}         
}
\toprule
Experiments & Num. & Pathological Objects\\
\midrule

Feature Importance Estimation & 12 & breast-cell-cancerous, breast-nuclei-ductal epithelium, breast-nuclei-epithelial, breast-cell-mitosis, breast-nuclei-neoplastic, breast-tissue-tumor, breast-tissue-stroma, breast-nuclei-fibroblast, breast-tissue-lymphocytic infiltrate, breast-tissue-necrosis or debris, breast-tissue-dcis, breast-nuclei-mitotic figure.\\
\midrule
Imaging Biomarker Discovery & 22 & breast-nuclei-epithelial, breast-tissue-tumor, breast-cell-glial, breast-tissue-dcis, breast-nuclei-neoplastic, breast-tissue-necrosis or debris, breast-cell-cancerous, breast-tissue-lymphocytic infiltrate, breast-nuclei-connective, breast-cell-mitosis, breast-tissue-stroma, breast-tissue-fat, breast-nuclei-lymphocyte, breast-nuclei-inflammatory, breast-tissue-plasma cells, breast-nuclei-neutrophil, breast-nuclei-fibroblast, breast-tissue-blood vessel, breast-nuclei-mitotic figure, breast-nuclei-ductal epithelium, breast-nuclei-plasma cells, other.\\
\bottomrule
\end{tabular}
\end{table*}





\end{appendices}

\end{document}